\documentclass[conference]{IEEEtran}
\PassOptionsToPackage{table}{xcolor}
\usepackage{times}

\usepackage[numbers]{natbib}
\usepackage{multicol}
\usepackage[bookmarks=true,hidelinks]{hyperref}

\usepackage{mymacros}
\usepackage{microtype}
\usepackage{graphicx}
\usepackage{booktabs}
\usepackage{algorithm2e}
\usepackage{color,soul}
\usepackage{lscape}
\usepackage{caption}
\usepackage{subcaption}
\usepackage{comment}
\usepackage{tabularx}
\usepackage{arydshln}
\usepackage{amsmath}
\usepackage{amssymb}
\usepackage{mathtools}
\usepackage{multirow}
\usepackage{amsthm}
\usepackage{bbm}
\usepackage{color,soul}
\usepackage{dsfont}
\usepackage{makecell} 
\usepackage{pifont}
\usepackage{siunitx}
\usepackage{placeins}
\usepackage{subcaption}
\usepackage[capitalize,noabbrev]{cleveref}
\pgfplotsset{compat=1.18}
\newcommand{\cmark}{\textcolor{brightGreen}{\ding{51}}}
\newcommand{\xmark}{\textcolor{brightRed}{\ding{55}}}

\begin{document}

\setlength{\textfloatsep}{8pt}

\title{Mind Your Steps: A General Learning Framework for Accurate Humanoid Foothold Tracking}

\author{
\authorblockN{Alessandro Montenegro\textsuperscript{1},\;
Shihao Li\textsuperscript{2},\;
Puze Liu\textsuperscript{2, 3, $\dagger$},\;
Alberto Maria Metelli\textsuperscript{1},\;and\;
Jan Peters\textsuperscript{3--6}}
\authorblockA{%
  \textsuperscript{1}Politecnico di Milano \hspace{0.1cm}
  \textsuperscript{2}Tongji University \hspace{0.1cm}
  \textsuperscript{3}Technische Universit\"{a}t Darmstadt \hspace{0.1cm}
  \textsuperscript{4}hessian.AI \\[0.3em]
  \textsuperscript{5}German Research Center for Artificial Intelligence (DFKI) \hspace{0.1cm}
  \textsuperscript{6}Robotics Institute Germany (RIG) \\[0.3em]
  {\small Project Website: \href{https://montenegroalessandro.github.io/mind-your-steps/}{\textcolor{gblue}{\textbf{\texttt{https://montenegroalessandro.github.io/mind-your-steps/}}}}}
}
}

\maketitle

\newcommand\blfootnote[1]{%
  \begingroup
  \renewcommand\thefootnote{}\footnote{#1}%
  \addtocounter{footnote}{-1}%
  \endgroup
}
\blfootnote{$^\dagger$Correspondence to: Puze Liu (\texttt{puze\_liu@tongji.edu.cn}).}

\begin{figure*}[t!]
    \centering
    \resizebox{.8\linewidth}{!}{\includegraphics[]{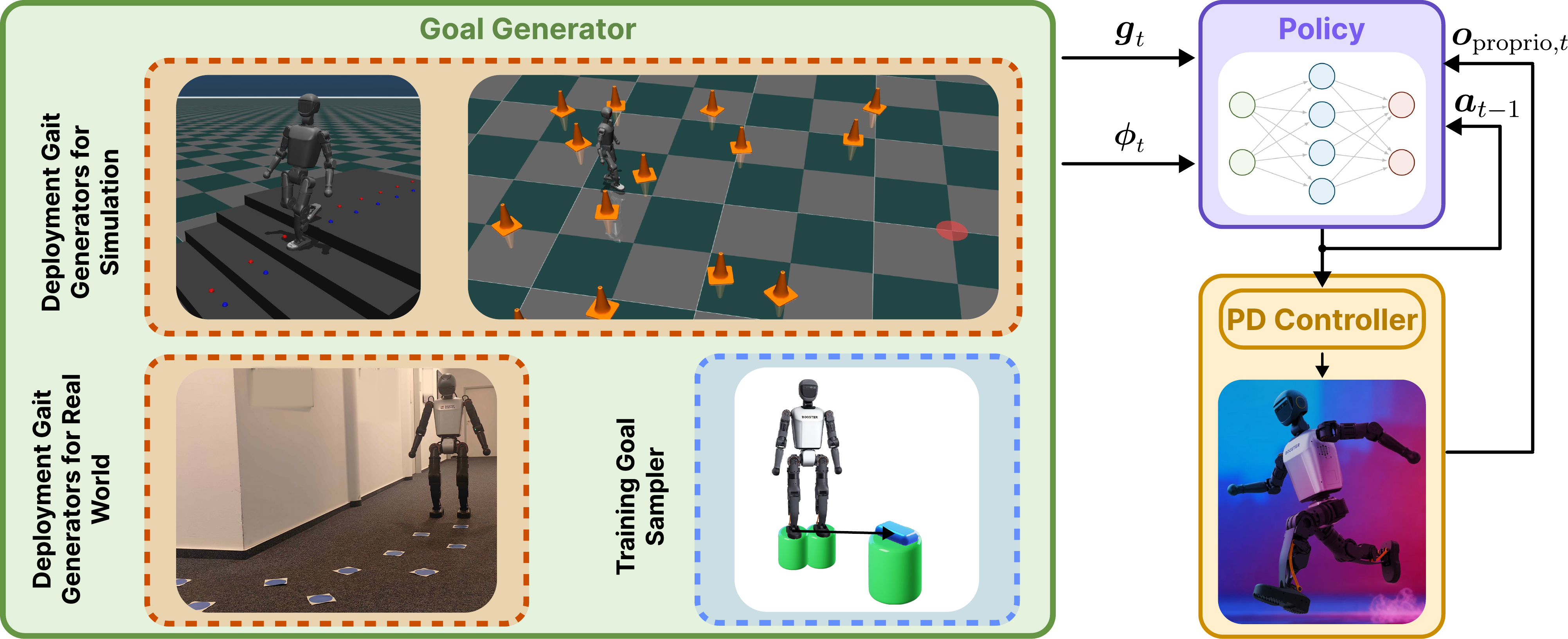}}
    \caption{\textbf{Training and Deployment Architecture.} Overview of the training and deployment architecture. \textit{Policy.} The policy $\pi_{\vtheta}$ receives proprioception $\bm{o}_{\text{proprio},t}$, the previous action $\bm{a}_{t-1}$, the foothold goal $\bm{g}_{t}$, and the gait phase $\bm{\phi}_{t}$. It outputs joint position targets for an underlying PD controller.
    \textit{Goal Generator.} This modular component supplies foothold targets to the policy. During \emph{training}, a procedural Goal Sampler generates synthetic targets to enforce tracking accuracy. During \emph{deployment} (simulation or real-world), this is replaced by task-specific planners, such as a vision-based estimator extracting visual targets from the ground or a teleoperation interface defining step direction, length, and yaw orientation.}
    \label{fig:train-arch}
\end{figure*}

\begin{abstract}
Enabling humanoid robots to operate in complex, dynamic environments remains a critical challenge, fundamentally limited by the ability to navigate robustly, safely, and accurately. 
While reinforcement learning with velocity-commanded policies has achieved remarkable robustness in humanoid locomotion, this approach lacks explicit control of the foothold placement, leading to unsafe behavior, such as stepping onto human feet, or imprecise navigation, hindering the following manipulation task.
Conversely, explicit foothold-tracking policies offer a promising alternative by directly being commanded with target foot poses. However, existing approaches are often limited by unrealistic state assumptions, compromising real-world deployment, or they are part of staged pipelines, making them tied to specific downstream tasks. 
In this work, we introduce a novel, lightweight framework for training general-purpose 3D foothold-tracking policies. By dynamically providing footstep support through a goal sampler, this method enables the learned policy to be agnostic to specific terrains. 
Our new target representation effectively mitigates challenges arising in the real world, such as noisy and inaccurate pose estimation and foot contact estimation.
Designed for direct real-world transfer, our policy acts as a standalone low-level controller that can be seamlessly paired with various high-level foothold generators. We demonstrate the effectiveness of our framework through extensive experiments in simulation and in the real world. By coupling our policy with different upstream planners, we achieve natural and accurate locomotion in challenging settings, paving the way for loco-manipulation tasks in complex environments.
\end{abstract}

\IEEEpeerreviewmaketitle

\section{Introduction} \label{sec:intro}

Recent research on humanoid robotics has demonstrated remarkable progress, enabling humanoids to perform agile and highly dynamic behaviors such as backflips and flying kicks. Despite these achievements, enabling humanoids to operate autonomously in real-world environments remains a critical challenge.  A fundamental prerequisite for real-world deployment is the ability to navigate safely, accurately, and reliably in everyday settings~\citep{kajita2003biped,pratt2006capture,bao2024deep,he2025attention}. 
In realistic environments, humanoid robots must contend with numerous challenges, including obstacle avoidance, interaction with humans, traversal of slopes and stairs, and locomotion over uneven or irregular terrain~\citep{zhang2024learning,huang2025traversing,wang2025beamdojo}. In such conditions, stable locomotion critically depends on precise and informed foot placement, as feasible footholds are often sparse and constrained. Consequently, endowing humanoids with controllers capable of constraining foot placements to valid and safe contact regions explicitly is a critical problem. 

Given its fundamental importance, legged locomotion has become one of the most studied fields in robotics. Extensive investigation, spanning model-based~\citep{pratt2006capture,winkler2018gait,jenelten2022tamols}, data-driven~\citep{lee2020learning,kumar2021rma,miki2022learning,long2023hybrid,nahrendra2023dreamwaq,zhuang2023robot,cheng2024extreme,he2024agile,margolis2024rapid,bao2024deep}, and hybrid methodologies~\citep{yu2021visual,gangapurwala2022rloc,xie2022glide,omar2023safesteps,jenelten2024dtc,huang2025traversing}, has yielded approaches with remarkable robustness in both simulation and real-world deployment. However, the predominant paradigm remains the integration of vision pipelines with velocity-commanded policies~\citep{wang2025beamdojo,huang2025traversing,he2025attention,ben2025gallant}. In this setting, the control policy typically relies on height map observations and auxiliary reward terms to implicitly select valid footholds while steering the robot away from unsteppable areas. However, the inherent mismatch between continuous velocity commands and discrete foot placement results in degraded tracking accuracy and reduced efficiency in reaching the target.

Since the aforementioned methodologies do not allow for direct control over the robot’s foot placement, explicit foothold-commanded policies have recently emerged as a promising alternative paradigm~\citep{peng2017deeploco,xie2020allsteps,duan2022sim,huang2025traversing,tsounis2020deepgait}.
These approaches are designed to enhance accuracy on foot placement targets and enable more flexible gait generation and motion capabilities. However, existing implementations remain severely limited in scope and practicality, far from general-purpose utility. Most have been studied solely in simulation~\citep{peng2017deeploco,xie2020allsteps}, exist only in preliminary forms restricted to simple forward walking on flat terrain~\citep{duan2022sim}, or are embedded as low-level components within specific, monolithic pipelines~\citep{peng2017deeploco,huang2025traversing}. Consequently, they often fail to support direct real-world usage and remain closely coupled to specific downstream tasks.

\textbf{Original Contribution.}~~Motivated by achieving stable and accurate locomotion via foothold-conditioned policies, in this work we propose a novel, lightweight, and model-free framework for training general-purpose 3D foothold-tracking policies. 
Such foothold-trackers are agnostic to the high-level planner and ready for immediate deployment, enabling seamless integration with diverse downstream tasks. Our specific contributions are as follows:
\begin{itemize}
    \item We propose a framework designed to train policies relying solely on proprioception, gait phase, and stance-relative foothold targets. By expressing commands in the stance-foot frame, the goal remains constant throughout the swing phase, removing the need for precise base state estimation~\cite{peng2017deeploco, xie2020allsteps}. To this end, we propose a novel Goal Sampler (GS) capable of generating feasible 3D foothold targets during training. Our GS does not tie with specific predefined terrains, enabling a general-purpose policy for downstream tasks. Coupled with specific reward terms balancing tracking accuracy with natural locomotion, this enables the agent to achieve accurate foot placement capabilities, thus generalizing locomotion in settings where real-world-like hurdles are present.
    \item Leveraging the independence of our foothold-tracking policy, we show its modularity by pairing it with various high-level planners in simulation. We validate performance on tasks such as goal reaching, cluttered environment navigation, and narrow paths and staircases traversal. In most scenarios, we compare against standard velocity-commanded policy, showing that our approach leads to more efficient and robust solutions.
    \item We deploy the foothold-tracking policy on the Booster T1 humanoid hardware. By integrating a simple vision-based foothold generator, we show zero-shot sim-to-real transfer, tracking the predefined foothold target successfully. This validates the framework's potential as a foundational block for complex loco-manipulation systems.
\end{itemize}
Figure~\ref{fig:train-arch} shows the proposed training and deployment pipeline.
\section{Related Work} \label{sec:related}

\textbf{Locomotion Over Sparse Footholds.}~~Locomotion over sparse footholds, where the main complexity arises from the combinatorial nature of foothold selection~\citep{corberes2025perceptive}, has been extensively studied as a perceptive locomotion problem using model-based methods~\citep{pratt2006capture,winkler2018gait,jenelten2022tamols}. Specifically, the literature has focused on developing hierarchical controllers relying on complex pipelines integrating perception, planning, and control~\citep{griffin2019footstep,jenelten2020perceptive,mastalli2020motion,melon2021receding,grandia2023perceptive,mishra2024efficient}. While years of research led to high-performing, lightweight, and interpretable approaches~\citep{pratt2006capture}, these methods often suffer from the common pitfalls of model-based control. Indeed, they are highly sensitive to model discrepancies and violations of modeling assumptions. Such problems are especially exacerbated in real world locomotion tasks due to the difficulty of terrain modeling, potentially resulting in unsafe or unreliable behavior during deployment.

To overcome the limitations of model-based methods, research has shifted toward model-free Reinforcement Learning~\citep[RL,][]{sutton1998reinforcement}, which learns control policies directly via interaction without relying on explicit priors. 
Leveraging domain randomization, these approaches have demonstrated remarkable robustness and generalization, enabling effective deployment on real-world hardware~\citep{lee2020learning,kumar2021rma,miki2022learning,long2023hybrid,nahrendra2023dreamwaq,zhuang2023robot,cheng2024extreme,he2024agile,margolis2024rapid,bao2024deep}. Typically, they learn 
velocity-tracking policies~\citep{agarwal2023legged,yang2023neural,cheng2024extreme,yu2024walking,zhang2024learning} paired with high-level vision components~\citep{he2025attention,ben2025gallant} whose integration is required in sparse foothold scenarios to identify valid places to step on.
A major issue in end-to-end model-free-based methodologies arises from the sparsity of valid footholds which results in sparse reward signals and frequent early episode terminations. To mitigate these limitations, the literature employed curriculum learning~\citep{narvekar2020curriculum} and proposed specialized reward structures and architectures, such as a reward account for foot geometry~\cite{wang2025beamdojo} or an attention-based terrain map~\cite{he2025attention} to achieve robust terrain-aware locomotion.

Another approach to addressing the limitations of both model-based and end-to-end model-free methodologies relies on hybrid architectures that combine the two~\citep{yu2021visual,gangapurwala2022rloc,xie2022glide,omar2023safesteps,jenelten2024dtc,huang2025traversing}. The objective is to train agents starting from physics priors~\cite{lee2024integrating}, aiming to retain the best of both worlds: high interpretability, robustness, and generalization capabilities. Model-based and model-free techniques can be merged in various ways. For instance, RL has been employed to generate trajectories that are subsequently tracked by model-based controllers~\citep{yu2021visual,gangapurwala2022rloc,xie2022glide}, or conversely, to train policies capable of tracking model-based planners~\citep{jenelten2024dtc,huang2025traversing}. 
While hybrid approaches preserve the benefits of both paradigms, they also inherit their respective pitfalls. Specifically, they rely on models that may fail when discrepancies with reality are significant. Furthermore, they can be computationally demanding and often require substantial implementation effort, particularly when extending established open-source robot locomotion frameworks~\citep[\eg][]{alhafez2023locomujoco}.

\textbf{Learning Foothold-Tracking Controllers.}~~Learning foothold-tracking controllers has gained significant attention in recent years. 
In most cases, such controllers function as part of two-stage architectures, serving as a bridge between high-level vision pipelines and underlying PD controllers to achieve accurate locomotion on sparse footholds.
However, these approaches often rely on unrealistic modeling assumptions, making them unsuitable for real-world deployment.
For instance, \citet{peng2017deeploco} proposed \emph{DeepLoco}, a two-stage architecture for terrain-aware humanoid locomotion employed solely in simulation. 
Besides requiring proprioceptive information and a gait phase indicator~\citep{siekmann2021sim} as input, the tracker additionally requires observing binary ground contact flags which are difficult to estimate accurately in the real world. 
Furthermore, the foothold goals consist of target positions for the next two steps~\citep{zaytsev2015two} and a target root heading, all expressed in the root frame, as also done in~\citep{singh2022learning}. This implies that real-world deployment would require precise state estimation/localization. Moreover, during training, foothold targets are extracted from motion capture clips to enforce stylistic mimicry. Despite these limitations, \citep{peng2017deeploco} is considered a seminal work, inspiring numerous follow-up studies~\citep{tsounis2020deepgait,jenelten2024dtc,he2025attention}.
Another key study focusing on simulation-based foothold tracking is \citep{xie2020allsteps}. The authors demonstrate that employing a carefully designed curriculum and a lightweight foothold sampler facilitates learning a tracker capable of traversing stepping stones and generalizing to continuous terrains. 
As with \citep{peng2017deeploco}, the reliance on binary contact states and body-frame targets renders such controllers difficult to deploy in the real world, likely necessitating specific high-level wrappers.
Moving toward real-world applicability, \citet{duan2022sim} extended the work of \citet{xie2020allsteps}. 
During training, they synchronize touchdown events with the gait clock and focus on forward motion on flat terrain. While simple, this approach paves the way for more sophisticated approaches capable of navigating real-world terrains for general-purpose tasks.
More recently, \citet{huang2025traversing} proposed an architecture specifically designed for traversing narrow paths in the real world, leveraging a Linear Inverted Pendulum (LIP) model~\citep{kajita2003biped}.
Then, they integrate a lightweight LiDAR-based perception module to refine the LIP-proposed footholds~\citep{silver2018residual,johannink2019residual}.

While foothold trackers have proven to be valuable for achieving robust and accurate locomotion by enabling full control over foot placement, to the best of our knowledge, they have been employed primarily as subordinate components within specific high-level pipelines. Furthermore, they often lack generality regarding downstream tasks or upstream gait generators and frequently face challenges in real-world deployment due to the reliance on specific, hard-to-estimate information in their observation space.
\section{Problem Formulation} \label{sec:formulation}


In RL, the environment is modeled as a Markov Decision Process~\citep[MDP,][]{puterman2014markov}. An MDP is represented by $\mathcal{M} \coloneqq \left( \mathcal{S}, \mathcal{A}, p, r, \rho_0, \gamma, T \right)$, where $\mathcal{S} \subseteq \mathbb{R}^{d_{\cS}}$ and $\mathcal{A} \subseteq \mathbb{R}^{d_{\mathcal{A}}}$ are the measurable state and action spaces; $p: \mathcal{S} \times \mathcal{A} \xrightarrow[]{} \Delta(\mathcal{S})$\footnote{$\Delta(\mathcal{X})$ is the set of probability distributions over a measurable space $\mathcal{X}$.} is the transition model, where $p(\bm{s}' | \bm{s}, \bm{a})$ is the probability density of landing in state $\bm{s}'\in \mathcal{S}$ by playing action $\bm{a}\in \mathcal{A}$ in state $\bm{s}\in \mathcal{S}$; $r: \mathcal{S} \times \mathcal{A} \xrightarrow[]{} [-R_{\max}, R_{\max}]$ is the reward function, where $r(\bm{s}, \bm{a})$ is the reward the agent gets when playing action $\bm{a}$ in state $\bm{s}$; $\rho_0 \in \Delta(\mathcal{S})$ is the initial-state distribution; $\gamma \in [0, 1]$ is the discount factor. A trajectory $\tau = \left( \bm{s}_{\tau,1}, \bm{a}_{\tau,1},\dots, \bm{s}_{\tau,T}, \bm{a}_{\tau,T} \right)$ of length $T \in \mathbb{N} \cup \{+\infty\}$ is a sequence of $T$ state-action pairs. In the following, we refer to $\mathcal{T}$ as the set of all trajectories. The \emph{discounted return} of a trajectory $\tau \in \mathcal{T}$ is $R(\tau) \coloneqq \sum_{t=1}^{T} \gamma^{t-1} r(\bm{s}_{\tau,t}, \bm{a}_{\tau,t})$. \footnote{We let $\gamma=1$ only when $T<+\infty$.} 

Most of the times, especially in robotics, at the interaction instant $t$ the agent is not provided with the full state $\bm{s}_{t}$, but just with a \emph{partial observation} of it. Specifically, it is provided with $\bm{o}_{t} \in \cO \subseteq \mathbb{R}^{d_{\cO}}$, which can be expressed as $\bm{o}_{t} = f(\bm{s}_{t})$, where $f: \mathcal{S} \to \mathcal{O}$.
Furthermore, as commonly done in this setting~\citep{wang2025beamdojo,huang2025traversing,jenelten2024dtc,he2025attention}, we frame our problem as a goal-based RL one, meaning that we extend the \emph{observation space} to incorporate a goal set $\cG \subseteq \mathbb{R}^{d_{\cG}}$, part of the state space of the MDP. Thus, the goal $\bm{g}_{t} \in \mathcal{G}$ influences the reward function $r$.

An RL agent interacts with the environment according to a \emph{parametric stochastic policy} $\pi_{\bm{\zeta}}: \cO \times \cG \to \Delta(\cA)$, where $\bm{\zeta} \in \mathcal{Z}$ is the policy parameterization in the parameter space $\mathcal{Z} \subseteq \mathbb{R}^{d_{\mathcal{Z}}}$.
The policy is employed to sample actions $\bm{a}_{t} \sim \pi_{\bm{\zeta}}(\cdot \mid \bm{o}_{t}, \bm{g}_{t})$ to be played at \emph{every interaction} instant $t$, being conditioned on the observation $\bm{o}_{t} \in \cO$ and the goal $\bm{g}_{t} \in \cG$.
The performance of a policy $\pi_{\vtheta}$ is assessed via the \emph{expected return} $J: \mathcal{Z} \to \mathbb{R}$, which is defined as: 
$J(\bm{\zeta}) \coloneqq \E_{\tau \sim p_{\bm{\zeta}}}\left[ R(\tau) \right]$,
where $p_{\bm{\zeta}}(\tau)$
is the density function of a trajectory $\tau$ induced by $\pi_{\bm{\zeta}}$. 
The goal is to learn the optimal policy $\pi_{\bm{\zeta}^{\star}}$, with $\bm{\zeta}^{\star} \in \argmax_{\bm{\zeta} \in \mathcal{Z}}J(\bm{\zeta})$.




\section{Methodology} \label{sec:method}
In this section, we detail the proposed framework. First, we define the foothold command formulation provided as input to the policy and describe the observation space, which is explicitly designed to facilitate real-world deployment. Next, we outline the characteristics of our training pipeline, detailing the employed agent structure and the goal sampler functioning. 
Finally, we introduce the reward function, focusing on the specific terms for foothold tracking and locomotion style. 

\subsection{Goal and Observation Formulation}
At each time step $t$, the policy observes \emph{proprioceptive information} $\bm{o}_{\text{proprio}}:=(\bm{\theta}_{\text{base}}, \bm{\omega}, \bm{q}, \bm{\dot{q}})$ (including base orientation $\bm{\theta}_{\text{base}}$, base angular velocity $\bm{\omega}$, joint positions $\bm{q}$ and velocities $\bm{\dot{q}}$) along with a \emph{goal command} specifying the desired footholds in terms of position and yaw orientation.
We formulate the foothold target goal as a vector $\bm{g}_{t}$, defined as:
\begin{align*}
\bm{g}_{t} \coloneqq \left( \prescript{s}{l}{\bm{p}}_{t}; \; \prescript{s}{l}{\bm{\psi}}_{t}; \; \prescript{s}{r}{\bm{p}}_{t}; \; \prescript{s}{r}{\bm{\psi}}_{t} \right),
\end{align*}
where $\prescript{s}{l}{\bm{p}}_{t}, \prescript{s}{r}{\bm{p}}_{t} \in \mathbb{R}^3$ denote the position offsets of the next desired left and right footholds, expressed in the \emph{current stance foot frame}. Similarly, $\prescript{s}{l}{\bm{\psi}}_{t}, \prescript{s}{r}{\bm{\psi}}_{t}$ represent the desired orientations (expressed as quaternions) in the \emph{stance foot frame}. As these foothold targets are expressed in the \emph{stance foot frame}, the targets are holding a constant value during each swing phase, significantly reducing target variability during movement compared to previous methods~\cite{peng2017deeploco,xie2020allsteps,singh2022learning}.

The agent also observes \emph{gait phase} information: $\bm{\phi}_{t} \coloneqq (\cos (2\pi\phi_{t}) ; \; \sin (2\pi\phi_{t}))$, where 
$\phi_{t} \in [0,1]$. If $\phi_{t} \in [0,0.5)$, the left foot acts as the swing foot; otherwise, the right foot swings.
As commonly done in this setting~\citep{peng2017deeploco,siekmann2021sim,xie2020allsteps}, we consider the phase variable to evolve like a clock, enforcing a constant locomotion cadence. Specifically, it evolves according to $\phi_{t+1} \leftarrow \phi_{t} + \delta_{\phi} \mod 1$, where $\delta_{\phi}$ is the \emph{fixed} gait offset determining the rapidity of gait phase switches.

\textbf{Design Choices.} Several aspects of this formulation differ from prior art. First, all targets in $\bm{g}_{t}$ are expressed in the \textit{stance-foot frame}. Consequently, the target remains constant until a phase switch occurs. Notice that the target corresponding to the current stance foot is effectively a zero offset (identity pose) for the whole phase duration. This contrasts with methods using the base frame~\citep{peng2017deeploco,xie2020allsteps,huang2025traversing,singh2022learning} or the specific foot frame~\citep{duan2022sim}. This choice facilitates real-world deployment by removing the need for precise base state estimation (localization) to update targets in real-time, as the relative command remains constant throughout the swing phase.
Second, we avoid providing binary ground contact flags~\citep{peng2017deeploco,xie2020allsteps}, which are difficult to estimate reliably on hardware.
Finally, by explicitly commanding 3D positions and yaw orientations, we ensure general-purpose navigation capabilities on complex terrain, surpassing prior works that control only subsets of these dimensions~\citep{duan2022sim,huang2025traversing}. Indeed, by training agents to track full 3D foothold targets, the humanoid enables omnidirectional locomotion (forward, backward, and lateral) while simultaneously controlling swing foot height and orientation. This comprehensive set of control possibilities allows the agent to execute precise footstep plans and complex walking patterns.

\subsection{Training Architecture and Goal Sampling} \label{subse:train}
For training our policy, we employ asymmetric Proximal Policy Optimization (PPO)~\citep{schulman2017proximal, pinto2018asymmetric}, a common used model-free RL algorithm in locomotion~\citep{bao2024deep}. We build upon the open-source library \emph{LocoMuJoCo}~\citep{alhafez2023locomujoco}, originally designed for velocity-commanded tasks, adapting its infrastructure to support our foothold-tracking framework.

\textbf{Agent Structure.}~~As done in asymmetric actor-critic, the actor $\pi_{\bm{\zeta}}$, which is parameterized by a $512 \times 256 \times 128$ network, receives proprioceptive observation as input, whereas the critic network shares the same network structure but takes privileged state as observation. The overall observation of the actor is 
\begin{align*}
    \bm{x}_{t} \coloneqq  \big( \bm{o}_{\text{proprio},t}; \bm{a}_{t-1}; \; \bm{\phi}_{t}; \; \bm{g}_{t} \big),
\end{align*}
where the components are defined in the preceding section. In addition to the actor's observation, the critic also observes the linear velocity of the robot base $\bm{v}_{\text{base}}$. 
Regarding the output action $\bm{a}_{t} \sim \pi_{\vtheta}(\cdot \mid \bm{x}_{t})$, we consider $\bm{a}_{t}$ to specify the target joint positions for a low-level PD controller, a common hierarchy in the literature~\citep{peng2017deeploco,huang2025traversing,he2025attention}. In this work, we target the Booster T1 humanoid, though the formulation is robot-agnostic. 
Figure~\ref{fig:train-arch} illustrates the complete architecture.

\begin{figure}[t]
    \centering
    
    \resizebox{.45\linewidth}{!}{\includegraphics{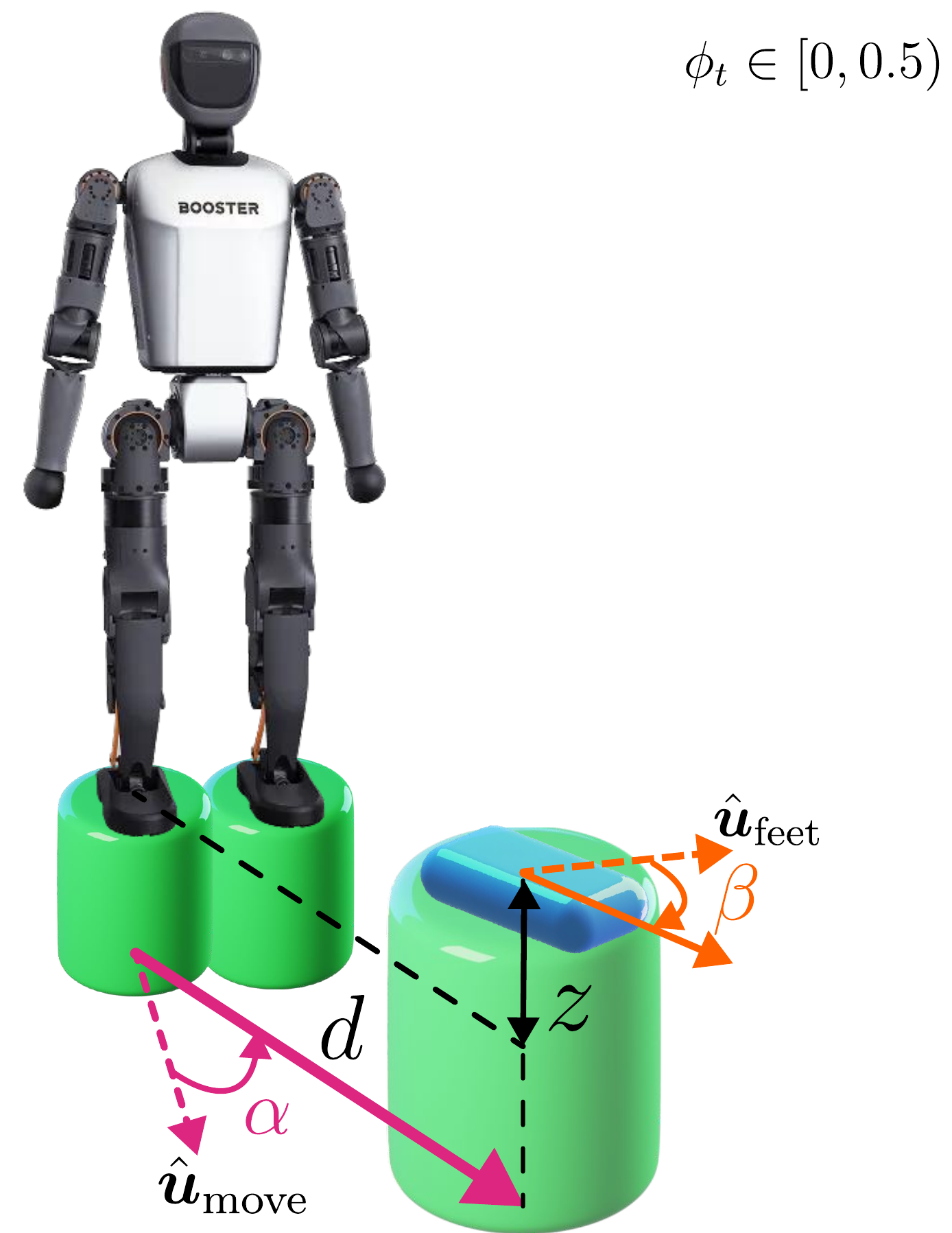}}
    
    \caption{\textbf{Goal Sampling.} Representation of the sampling process during a left-swing phase ($\phi_{t} < 0.5$). The global vectors $\hat{\bm{u}}_{\text{move}}$ and $\hat{\bm{u}}_{\text{feet}}$ define the nominal locomotion heading and foot orientation, fixed for the entire episode. At each gait switch, the angles $\alpha$ and $\beta$ are sampled from independent uniform distributions to perturb the nominal directions. The scalars $d$ and $z$ determine the planar distance and vertical offset, respectively, relative to the current stance foot.}
    \label{fig:3d-sampling}
\end{figure}

\textbf{Goal Sampling.}~~During training, we employ a Goal Sampler (GS) responsible for generating feasible foothold targets and supports. In the following, we denote quantities in the world frame $w$ unless otherwise specified. We remark that the GS operates just in simulation, while the agent is then provided with the observation described above.
At the \emph{beginning of each episode} ($t=1$), the GS defines the global locomotion heading. It samples a desired movement direction $\theta_{\text{move}} \in [-\pi, \pi)$ and a nominal feet orientation $\theta_{\text{feet}} \in [-\pi, \pi)$.
These global directions \emph{remain constant} throughout the episode to enforce motion consistency. The gait phase $\phi_{1}$, defining the start of the motion with either the left or the right foot, is initialized conditioned to $\theta_{\text{move}}$ or sampled at random whenever possible.
\emph{Target sampling} is triggered at every \emph{gait phase switch} (\ie whenever the swing and stance roles swap). 
The GS computes the target position $\prescript{w}{}{\bm{p}}_{\text{tgt}}$ and yaw orientation $\prescript{w}{}{\psi}_{\text{tgt}}$ for the swing foot as follows:
\begin{enumerate}
    \item The GS samples a deviation angle $\alpha \sim \mathcal{U}(\alpha_{\min}, \alpha_{\max})$\footnote{$\mathcal{U}(a,b)$ denotes a uniform distribution continuous in $[a,b]$.} and a step length $d \sim \mathcal{U}(d_{\min}, d_{\max})$. The planar displacement is computed by perturbing the nominal movement direction $\theta_{\text{move}}$:
    \begin{align*}
        \Delta \bm{p} = \left( d \cos(\theta_{\text{move}} + \alpha) ; \; d \sin(\theta_{\text{move}} + \alpha); \; 0 \right).
    \end{align*}
    This ensures that while the robot follows a general direction, it must handle lateral and angular variations.
    \item Similarly, the target foot yaw is generated by perturbing the global feet orientation with an offset $\beta \sim \mathcal{U}(\beta_{\min}, \beta_{\max})$, yielding $\prescript{w}{}{\psi}_{\text{tgt}} = \theta_{\text{feet}} + \beta$.
    \item The vertical component is sampled as an offset $z \sim \mathcal{U}(z_{\min}, z_{\max})$ relative to the current stance foot height. The final world-frame target is:
    \begin{align*}
        \prescript{w}{}{\bm{p}}_{\text{tgt}} = \prescript{w}{s}{\bm{p}} + \Delta \bm{p} + \left(0;\;0;\;z\right).
    \end{align*}
    Crucially, to ensure physical feasibility in simulation, we implement a \textit{terrain adaptation} mechanism: the environment consists of moving pillars that automatically adjust their height to match the sampled $z$-coordinate of the target, guaranteeing a valid contact surface. 
\end{enumerate}
Finally, the sampled target $(\prescript{w}{}{\bm{p}}_{\text{tgt}}, \prescript{w}{}{\psi}_{\text{tgt}})$ is transformed into current stance foot frame and clipped to feasible region to avoid leg crossing $(\prescript{s}{\Box}{\bm{p}}_{t}, \prescript{s}{\Box}{\bm{\psi}_{t}})$ , where $\Box \in \{l,r\}$ depending on the current swing foot index, 
to build the goal vector $\bm{g}_{t}$ as previously described. We highlight that in $\bm{g}_{t}$ the part related to the stance foot is taken as the offsets in the stance foot frame of the last observed position and orientation of the stance foot itself at the \emph{gait switch instant}.
To enable \emph{standing capabilities}, with probability $p_{\text{hold}}$, the GS enters a \quotes{hold-still} mode. In this case, the target position is set to maintain a nominal stance width relative to the stance foot, the orientation is aligned with the current stance foot (identity relative quaternion), and the gait phase is frozen providing the agent with $\bm{\phi}_{t} = (0; 0)$ for every $t$ in which the GS remains in the still mode.
Figure~\ref{fig:3d-sampling} shows the sampling strategy.

\textbf{Additional Training Details.}~~As previously mentioned, we utilize Asymmetric PPO~\citep{schulman2017proximal,pinto2018asymmetric}, a standard model-free RL algorithm in robot learning~\citep{bao2024deep}. Specifically, we adopt an asymmetric actor-critic architecture, where the critic has access to privileged simulation states unavailable to the actor.
We define an episode as \emph{terminated} (absorbing state) if the robot's base height falls below a critical threshold.
Furthermore, we employ extensive \emph{domain randomization}, including perturbations to link geometries, masses, and contact friction, as well as observation noise and external force disturbances (simulating kicks). Further details are provided in the supplementary material.
Finally, we implement a \emph{two-stage training} curriculum. Initially, we train the agent on flat terrain ($z_{\min}=z_{\max}=0$) to learn a robust planar tracker, then we fine-tune the agent on full 3D terrain.

\subsection{Reward Design}~~The reward function is mainly composed by tracking reward terms, the responsible ones for making the agent learn foothold-tracking capabilities and penalization terms employed to enforce a natural locomotion style. Penalization terms, which are taken from the velocity-based locomotion reward implemented in \emph{LocoMuJoCo}~\citep{alhafez2023locomujoco}, allow to learn a natural locomotion style, without relying on motion capture-based terms~\citep{peng2017deeploco}. In this section we focus on tracking terms, while the penalization ones are detailed in appendix.

\textbf{Foothold-Tracking Terms.}~~The fundamental term of our reward function is represented by the one responsible for tracking the goal $\bm{g}_{t}$. Let $(\prescript{w}{\Box}{\bar{\bm{p}}}_{t}; \prescript{w}{\Box}{\bar{\bm{\psi}}}_{t})$, with $\Box \in \{l,r\}$, be the target foothold positions and orientations to be tracked by both feet. We define the foothold-tracking term as follows:
\begin{align*}
    r_{\text{track}}(\bm{s}_{t}, \bm{a}_{t}) = r_{\text{sw}}(\bm{s}_{t}, \bm{a}_{t}) + r_{\text{st}}(\bm{s}_{t}, \bm{a}_{t}),
\end{align*}
where $r_{\text{sw}}$ and $ r_{\text{st}}$ refer to the swing and stance foot, respectively.
Notice that we explicited the fact that this reward term depends on the state $\bm{s}_{t}$ rather than on the observation $\bm{o}_{t}$, since it needs world frame coordinates, always accessible in simulation. Specifically, we define \emph{swing}-related term as:
\begin{align*}
    r_{\text{sw}}(\bm{s}_{t}, \bm{a}_{t}) &= \omega_{1} \exp \left( -\xi_{1} \| \prescript{w}{\Box}{\bm{p}}_{t,xy} - \prescript{w}{\Box}{\bar{\bm{p}}}_{t,xy} \|_{2}^{2} \right) \\
    &\quad + \omega_{2} \exp \left( -\xi_{2} ( \prescript{w}{\Box}{\bm{p}}_{t,z} - \prescript{w}{\Box}{\bar{\bm{p}}}_{t,z} )^{2} \right) \\
    &\quad + \omega_{3} \exp\left( -\xi_{3} \left( \prescript{w}{\Box}{\psi}_{t} - \prescript{w}{\Box}{\bar{\psi}}_{t} \right)^{2}\right),
\end{align*}
where $\Box \in \{l,r\}$ depends on the current $\phi_{t}$ value, $\prescript{w}{\Box}{\psi}_{t}$ and $\prescript{w}{\Box}{\bar{\psi}}_{t}$ are the yaws (wrapped in $[-\pi,\pi)$) relative to the quaternions $\prescript{w}{\Box}{\bm{\psi}}_{t}$ and $\prescript{w}{\Box}{\bar{\bm{\psi}}}_{t}$, and $\bm{p}_{t,xy}$ and $\bm{p}_{t,z}$ represent the planar and the height components of vector $\bm{p}_{t}$, respectively. 
The $r_{\text{sw}}(\bm{s}_{t}, \bm{a}_{t})$ explicitly allows the policy to track the 3D target foothold for the swinging foot. Notice that decoupling planar and height components in $r_{\text{sw}}(\bm{s}_{t}, 
\bm{a}_{t})$ allows to define different sharpness coefficients $\xi_{1}$ and $\xi_{2}$. In this context, this is needed given the different ranges covered by planar and height components.
Finally, the \emph{stance} reward term $r_{\text{st}}(\bm{s}_{t}, \bm{a}_{t})$ is designed to retain the final value of the swing reward $r_{\text{sw}}$ computed at the moment of touchdown. Specifically, for the entire duration of the stance phase, this term equals the accuracy achieved by the foot at the instant it transitioned from swing to stance.

\begin{figure}
    \centering
    \resizebox{\linewidth}{!}{\includegraphics[]{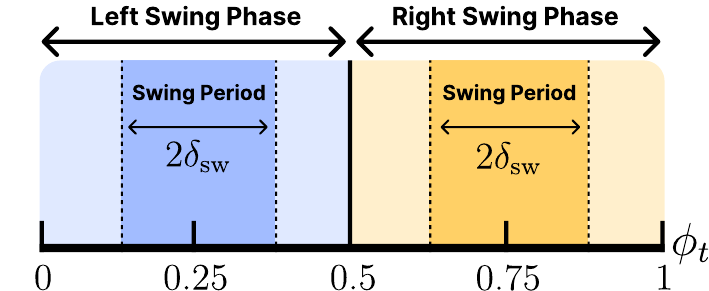}}
    \caption{\textbf{Swing-based Reward.} Illustration of the desired swing interval. The reward term $r_{\text{feet}}$ incentivizes foot clearance specifically when the phase variable $\phi_{t}$ lies within the window $2\delta_{\text{sw}}$. A non-zero reward is accumulated only if the correct foot breaks contact with the ground during this allotted time.}
    \label{fig:swing}
\end{figure}

\textbf{Feet Swing Term.}~~To ensure adherence to the commanded gait schedule, we introduce the term $r_{\text{feet}}$. This reward encourages the feet to lift during the allotted swing phase defined by the variable $\phi_{t}$. Similar swing-phase incentives are standard in humanoid locomotion frameworks~\citep{alhafez2023locomujoco}. As depicted in Figure~\ref{fig:swing}, we employ a temporal window within each gait phase where foot clearance is mandatory.
Formally:
\begin{align*}
    r_{\text{feet}}(\bm{s}_{t}, \bm{a}_{t}) &= \omega_{\text{feet}} \mathds{1}\left\{ | \phi_{t} - 0.25 | \le \delta_{\text{sw}}  \right\} \cdot \mathds{C}_{l} \\
    &\quad + \omega_{\text{feet}} \mathds{1}\left\{ | \phi_{t} - 0.75 | \le \delta_{\text{sw}}  \right\} \cdot \mathds{C}_{r},
\end{align*}
where $\delta_{\text{sw}}$ represents the half-width of the swing window, and $\mathds{C}_{\Box}$ ($\Box \in \{l,r\}$) acts as a binary \quotes{no-contact} indicator (returning $1$ when the foot is in the air).
Also in this case, this term depends on the full state $\bm{s}_{t}$ rather than the observation $\bm{o}_{t}$, as it relies on ground truth contact data. Note that, unlike the tracking term, $r_{\text{feet}}$ is purely locomotion-related and does not depend on the foothold goal $\bm{g}_{t}$.

\textbf{Swing Knee Height Term.}~~The last fundamental tracking term encourages the robot to raise the knee of the swing leg, a particularly important skill when tracking the $z$ component of the desired foothold. Specifically, lifting the knee ensures sufficient clearance to bring the swing foot to the target height, avoiding collisions with steps or obstacles. This term is defined exclusively for the swing leg.
The reward term is as follows:
\begin{align*}
    \resizebox{\linewidth}{!}{$\displaystyle r_{\mathrm{k}}(\bm{s}_{t},\bm{a}_{t}) = \omega_{\mathrm{k}} \exp \left( -\xi_{\mathrm{k}} \max\left( \prescript{w}{\Box}{\bm{\overbar{p}}}_{t,z} + \delta_{\mathrm{k}} - \prescript{w}{\Box,\mathrm{k}}{\bm{p}}_{t,z}, \, 0 \right)^{2} \right),$}
\end{align*}
where $\Box \in \{\mathrm{l},\mathrm{r}\}$ depends on the current $\phi_{t}$ to select the swing leg, $\prescript{w}{\Box}{\bm{\overbar{p}}}_{t,z}$ is the $z$ target for the swing foot in the world frame, $\delta_{\mathrm{k}}$ is a slack representing a knee clearance margin, and $\prescript{w}{\Box,\mathrm{k}}{\bm{p}}_{t,z}$ is the world-frame $z$ coordinate of the knee joint of the swing leg. This term is one-sided: the reward is maximal whenever the knee exceeds the threshold $\prescript{w}{\Box}{\bm{\overbar{p}}}_{t,z} + \delta_{\mathrm{k}}$, and decays otherwise. 

\begin{figure}[t]
    \centering
    \resizebox{\linewidth}{!}{\includegraphics[]{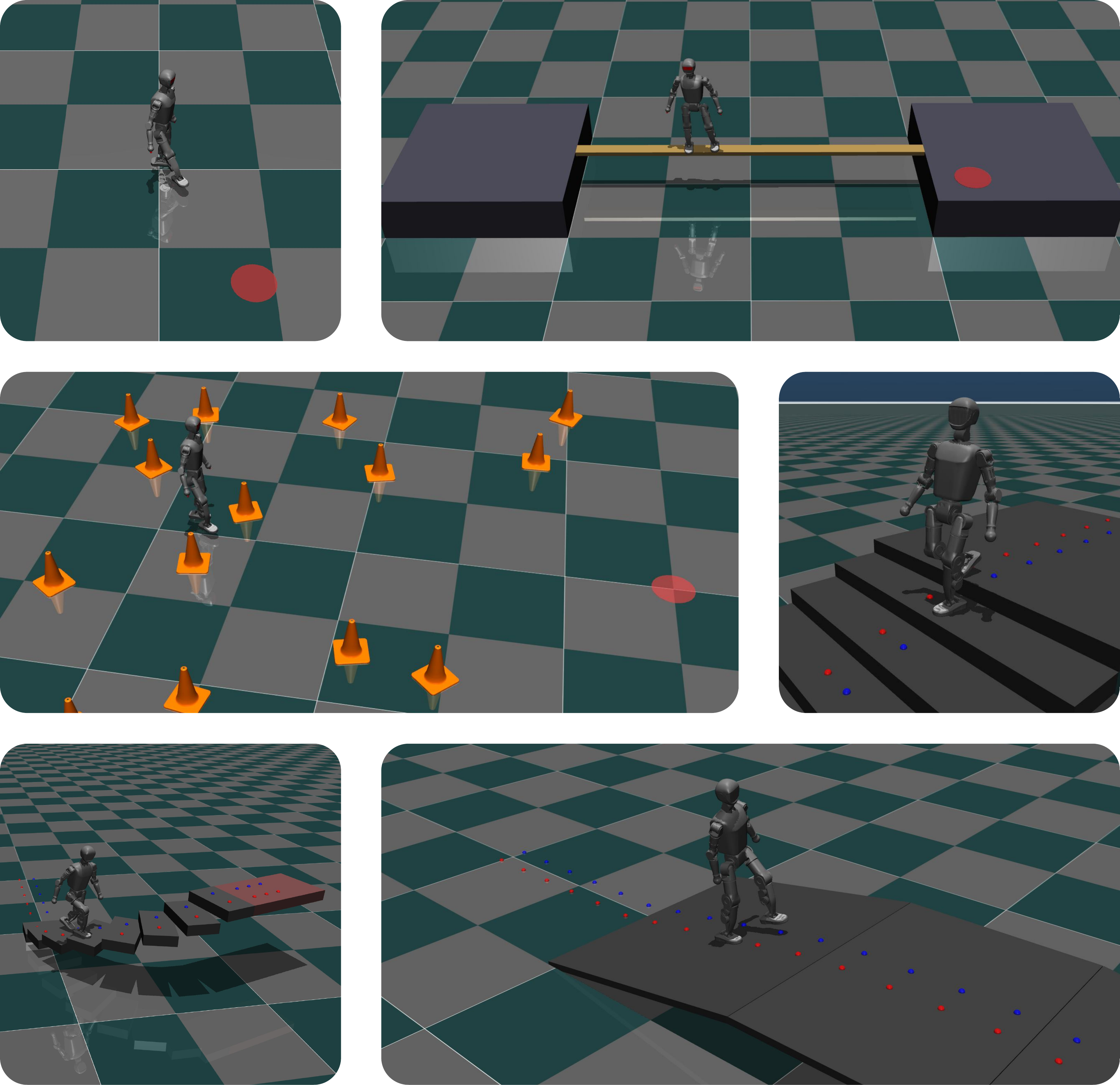}}
    \caption{\textbf{Evaluation Tasks in Simulation.} Simulation environments considered in Section~\ref{sec:exp-sim}. Evaluations were conducted on: goal reaching, narrow bridge traversal, navigation in cluttered environments, and stairs and ramp climbing.}
    \label{fig:tasks}
\end{figure}

\section{Experiments} \label{sec:exp-sim}
We evaluate the effectiveness of the proposed framework in simulation by comparing our Foothold-Tracking (FT) policy ($\pi_{\text{FT}}$) against a state-of-the-art Velocity-Tracking (VT) baseline ($\pi_{\text{VT}}$) trained via \emph{LocoMuJoCo}~\citep{alhafez2023locomujoco}. 
Due to the discrepancy of the command types, to ensure a fair comparison, both policies are trained within a equivalent command region, where velocity command $v_{\text{FT}}$ can be converted by the foot distance $v_{\text{VT}}$ and frequency $\nu$ as $d_{\text{FT}}$ $v_{\text{VT}}= 2d_{\text{FT}} \nu$.
Through a comprehensive set of carefully designed experiments, we demonstrate the agility, accuracy, and flexibility of our FT policy, showing that it enables a general-purpose locomotion policy with a wide range of motion capabilities. We use multiple task-specific high-level planners for $\pi_{\text{FT}}$ in each experiment. This highlights the framework's modularity, demonstrating that the learned policy serves as a general-purpose low-level controller capable of seamless integration with diverse upstream generators. Evaluation environments are illustrated in Figure~\ref{fig:tasks}.

\subsection{Goal Reaching Task}

\begin{figure}[t]
    \centering
    \includegraphics[width=0.9\linewidth]{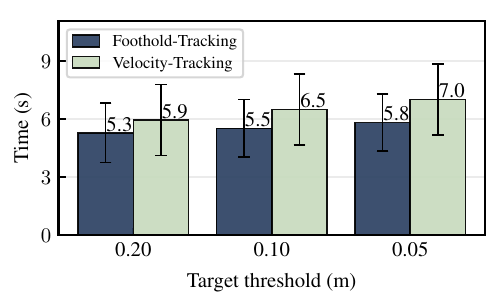}
    \caption{\textbf{Goal-Reaching Experiment.} Time required to reach a target below different thresholds. Mean mean and standard deviation are shown for each setting with 1000 trials.}
    \label{fig:goal-reaching-perf}
\end{figure}

We first evaluate the policies on the task of reaching a goal precisely and efficiently, which is an essential requirement for whole-body manipulation tasks.
Here, we deploy a variant of $\pi_{\text{FT}}$ trained on flat ground to isolate planar maneuvering performance.
Targets are sampled in a $[-3, 3]$m region, and task success is defined by multiple distance-based thresholds.
Additionally, we compare $\pi_{\text{FT}}$ against $\pi_{\text{FT-R}}$, a baseline identical to our method except that foothold targets are expressed in the \emph{root} frame rather than the \emph{stance foot} frame. This ablation isolates the contribution of the stance-foot frame formulation.

\textbf{High-Level Planners.}~~For $\pi_{\text{FT}}$ and $\pi_{\text{FT-R}}$, we implement a simple planner that generates footholds along the vector connecting the robot's base position to the goal. The step length is clamped to a maximum of $0.5$ m, and footholds are constrained to feasible regions to prevent leg crossing. In the meantime, the foot orientation is steered towards the target direction to encourage forward walking. Footholds are generated respecting the gait frequency the policy saw during training. 
For $\pi_{\text{VT}}$, we employ a standard proportional controller: the target velocity is computed as $\bm{v}_{\text{cmd}} = K (\bm{p}_{\text{goal}} - \bm{p}_{\text{base}})$, saturated to the robot's maximum velocity limits. The angular velocity is controlled by a separate P-controller to maintain the heading towards the target. In both methods, we assume the base positions are directly observable. Although unrealistic for real-world deployments, it allows us to decouple the effect from the localization errors encountered by both methods.

\textbf{FT vs. VT.}~~We quantify performance by measuring the time-to-goal under three distinct acceptance radii, specifically, $0.2$m, $0.1$m, and $0.05$m. For each threshold, we evaluate 1,000 trials with randomly sampled target positions.
As shown in Figure~\ref{fig:goal-reaching-perf}, $\pi_{\text{FT}}$ consistently outperforms $\pi_{\text{VT}}$, reaching the target significantly faster across all acceptance radii. This performance gain exhibit higher agility of the foot-tracking policy due to explicit control of the feet. 
Notably, as the success threshold decreases, the time-to-goal increases for $\pi_{\text{FT}}$ by only 5.3\%, whereas $\pi_{\text{VT}}$ exhibits a substantially larger increase of 18.9\%. This degradation for $\pi_{\text{FT}}$ majorly stems from the inherit mismatch between the continuous velocity commands and discretized footstep execution, which necessitates multiple correcting steps near the target. On the contrary, foothold-tracking directly command next foothold based on its current state and is therefore limited mainly by tracking accuracy. 
The results validate that explicit foothold control enables more aggressive and precise maneuvering compared to velocity-based commands, further narrowing the gap between high-level planning and low-level execution.

\begin{table}[t]
    \centering
    \resizebox{\linewidth}{!}{%
    \begin{tabular}{ccccc}
        \toprule
        \makecell[c]{\textbf{Policy} \\ ($\sigma_{\text{xyz}}$ (m), $\sigma_{\text{yaw}}$ (deg))} & \makecell{FT (Ours) \\ ---} & \makecell{FT-R \\(0, 0)} & \makecell{FT-R \\($3e{-2}$, 2.0)} & \makecell{FT-R \\($6e{-2}$, 3.0)} \\
        \midrule
        \makecell[c]{Accuracy (cm)\\{\scriptsize (mean $\pm$ std})} & 2.15 $\pm$ 0.164 & 1.66 $\pm$ 0.170 & 2.07 $\pm$ 0.237 & 2.86 $\pm$ 0.264 \\
        \bottomrule
    \end{tabular}}
    \caption{\textbf{Frame Ablation.} Foothold-tracking accuracy across 1000 targets on flat terrain, comparing $\pi_{\mathrm{FT}}$ and $\pi_{\mathrm{FT-R}}$. $\sigma_{\text{xyz}}$~(m) and $\sigma_{\text{yaw}}$~(deg) are the position and yaw noise scales for estimating base estimation errors for $\pi_{\mathrm{FT-R}}$.
    }
    
    \label{tab:frame-ablation}
\end{table}

\textbf{Frame Ablation.}~~Table~\ref{tab:frame-ablation} reports foothold-tracking accuracy across 1000 targets on flat terrain. To simulate real-world deployment conditions, we inject uniform noise into the base state estimate of $\pi_{\mathrm{FT-R}}$, considering two noise scales to cover a range from mild to realistic localization error. Without noise, $\pi_{\mathrm{FT-R}}$ improves over $\pi_{\mathrm{FT}}$ in tracking accuracy, as the root frame provides a precise target signal under ideal conditions. However, performance degrades sharply even under small noise magnitudes, demonstrating that root-frame policies are sensitive to localization errors. By contrast, $\pi_{\mathrm{FT}}$ is unaffected by state estimation noise by design, validating the stance-foot frame as the more robust formulation for real-world deployment.

\subsection{Narrow Bridge Traversal}



In this experiment, we challenge motion accuracy between two tracking policies ($\pi_{\text{FT}}$ and $\pi_{\text{VT}}$) in a narrow bridge traversal task as shown in Fig~\ref{fig:tasks}-UpperRight. This scenario demands high-precision foot placement, as the tolerance for error is extremely low. The robot is initialized at a different location, aiming to pass a bridge with different widths and lengths.

\begin{figure}[tb!]
    \centering
    \includegraphics[width=\linewidth]{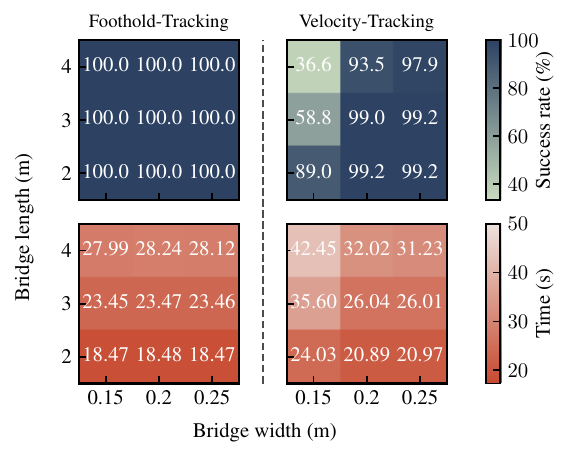}
    \caption{\textbf{Narrow Bridge Experiment.} The success rate($\uparrow$) and traversing time($\downarrow$) to cross a narrow bridge of different widths and lengths. Bridge width varies from 0.15 to 0.25m and length varies from 2 to 4m. Each configuration is conducted with 1000 trials with randomly sampled initial states and targets.}
    \label{fig:narrow-perf}
\end{figure}

\textbf{High-Level Planners.}~~Both agents employ similar planners introduced in the goal-reaching experiment, with additional task-specific modifications. In particular, auxiliary waypoints are placed at the entrance and exit of the bridge to enforce the robot moving parallel to the bridge's longitudinal direction. Due to the bridge’s restricted width, successful traversal requires precise lateral control. For $\pi_{\text{FT}}$, the planner generates foothold targets that explicitly maintain the robot’s footsteps near the centerline of the bridge. For $\pi_{\text{VT}}$, we increase the control gain on the robot's heading direction to correct deviations that could otherwise result in a fall. 

\textbf{Results.}~~We evaluate performance by measuring both success rate and traversal time across a range of bridge widths (0.15–0.25 m) and lengths (2–4 m). While both methods achieve high success rates, for less challenging configurations with a wider and shorter bridge, our $\pi_{\text{FT}}$ controller consistently completes the traversal in less time, indicating more efficient motion execution. As the task difficulty increases, in the critical scenarios with a width of 0.15m, narrower than the robot's foot length, the success rates for the VT controller drops drasticaly, whereas the FT-controller maintains almost the same performance. This is because the FT-tracking explicitly sets the desired target to be at the center of the bridge, eliminating the errors derived from proportional controller. 
Our foothold tracking controller provides a transparent and interpretable control interface enabling precise step-level reasoning and correction. This capability is crucial for safety-critical locomotion tasks, where accurate contact placement is required to prevent catastrophic failures, demonstrating the suitability of foothold-based control for real-world, constrained navigation scenarios.


\begin{table}[t]
\centering
\begin{tabular}{cccc}
\toprule
\# Cones & \textbf{Success (\%)} & Fall (\%) & Plan fail (\%) \\
\midrule
10 & 99 & 1 & 0 \\
15 & 98 & 0 & 2 \\
20 & 90 & 0 & 10  \\
\bottomrule
\end{tabular}
\caption{\textbf{Cluttered Environment Experiment.} 100 episodes in a $6\,\mathrm{m}\times6\,\mathrm{m}$ workspace with randomly placed cones.}
\label{tab:clutter_outcomes}
\end{table}

\subsection{Traversing Complex Environments}
Finally, we demonstrate the integration of the trained $\pi_{\mathrm{FT}}$ in four additional tasks: cluttered environment traversal, straight and spiral staircase climbing, and ramp traversal.

\begin{table*}[t]
    \centering
    \begin{subtable}[t]{0.49\linewidth}
        \centering
        \resizebox{\linewidth}{!}{{\setlength{\tabcolsep}{5pt}
\renewcommand{\arraystretch}{1.3}
\scriptsize
\begin{tabular}{c ccccccc}
    \toprule
    \multirow{2}{*}{\textbf{\makecell{Step\\Length\\(m)}}}
      & \multicolumn{7}{c}{\textbf{Step Height (m)}} \\
    \cmidrule(lr){2-8}
    & \textbf{0.08} & \textbf{0.10} & \textbf{0.11} & \textbf{0.115}
    & \textbf{0.12} & \textbf{0.125} & \textbf{0.13} \\
    \midrule
    
    \textbf{0.300} &
    \cellcolor{ggreen!20}\makecell{95\%\\R:3~~B:2\\0.059$\pm$0.089} &
    \cellcolor{ggreen!20}\makecell{76\%\\R:19~~B:5\\0.071$\pm$0.049} &
    \cellcolor{gyellow!20}\makecell{60\%\\R:40~~B:0\\0.081$\pm$0.048} &
    \cellcolor{gyellow!20}\makecell{52\%\\R:34~~B:14\\0.090$\pm$0.048} &
    \cellcolor{gyellow!20}\makecell{56\%\\R:39~~B:5\\0.083$\pm$0.041} &
    \cellcolor{gred!20}\makecell{38\%\\R:47~~B:15\\0.095$\pm$0.038} &
    \cellcolor{gred!20}\makecell{12\%\\R:70~~B:18\\0.108$\pm$0.039} \\
    
    \midrule
    \textbf{0.325} &
    \cellcolor{ggreen!20}\makecell{100\%\\R:0~~B:0\\0.038$\pm$0.005} &
    \cellcolor{ggreen!20}\makecell{100\%\\R:0~~B:0\\0.041$\pm$0.006} &
    \cellcolor{ggreen!20}\makecell{92\%\\R:5~~B:3\\0.050$\pm$0.029} &
    \cellcolor{ggreen!20}\makecell{89\%\\R:11~~B:0\\0.053$\pm$0.027} &
    \cellcolor{gyellow!20}\makecell{74\%\\R:22~~B:4\\0.068$\pm$0.042} &
    \cellcolor{gred!20}\makecell{30\%\\R:54~~B:16\\0.104$\pm$0.045} &
    \cellcolor{gred!20}\makecell{6\%\\R:80~~B:14\\0.115$\pm$0.038} \\
    
    \midrule
    \textbf{0.350} &
    \cellcolor{ggreen!20}\makecell{100\%\\R:0~~B:0\\0.040$\pm$0.005} &
    \cellcolor{ggreen!20}\makecell{99\%\\R:1~~B:0\\0.045$\pm$0.022} &
    \cellcolor{ggreen!20}\makecell{100\%\\R:0~~B:0\\0.045$\pm$0.016} &
    \cellcolor{ggreen!20}\makecell{96\%\\R:2~~B:2\\0.050$\pm$0.022} &
    \cellcolor{ggreen!20}\makecell{89\%\\R:7~~B:4\\0.066$\pm$0.056} &
    \cellcolor{gyellow!20}\makecell{47\%\\R:34~~B:19\\0.106$\pm$0.060} &
    \cellcolor{gred!20}\makecell{19\%\\R:58~~B:23\\0.136$\pm$0.057} \\
    
    \midrule
    \textbf{0.375} &
    \cellcolor{ggreen!20}\makecell{100\%\\R:0~~B:0\\0.048$\pm$0.011} &
    \cellcolor{ggreen!20}\makecell{100\%\\R:0~~B:0\\0.052$\pm$0.010} &
    \cellcolor{ggreen!20}\makecell{99\%\\R:0~~B:1\\0.054$\pm$0.017} &
    \cellcolor{ggreen!20}\makecell{86\%\\R:4~~B:10\\0.058$\pm$0.019} &
    \cellcolor{ggreen!20}\makecell{88\%\\R:8~~B:4\\0.079$\pm$0.100} &
    \cellcolor{ggreen!20}\makecell{87\%\\R:6~~B:7\\0.087$\pm$0.112} &
    \cellcolor{gyellow!20}\makecell{46\%\\R:23~~B:31\\0.113$\pm$0.086} \\
    
    \bottomrule
\end{tabular}}}
        \caption{\textbf{Straight Stairs.} }
        \label{tab:stairs}
    \end{subtable}
    \hfill
    \begin{subtable}[t]{0.49\linewidth}
        \centering
        \resizebox{\linewidth}{!}{{\setlength{\tabcolsep}{5pt}
\renewcommand{\arraystretch}{1.3}
\scriptsize
\begin{tabular}{c ccccccc}
\toprule
\multirow{2}{*}{\textbf{\makecell{Step\\Length\\(m)}}}
  & \multicolumn{7}{c}{\textbf{Step Height (m)}} \\
\cmidrule(lr){2-8}
& \textbf{0.08} & \textbf{0.10} & \textbf{0.11} & \textbf{0.115}
& \textbf{0.12} & \textbf{0.125} & \textbf{0.13} \\
\midrule

\textbf{0.300} &
\cellcolor{ggreen!20}\makecell{97\%\\R:2~~B:1\\0.045$\pm$0.017} &
\cellcolor{ggreen!20}\makecell{82\%\\R:17~~B:1\\0.061$\pm$0.029} &
\cellcolor{gyellow!20}\makecell{40\%\\R:40~~B:20\\0.082$\pm$0.028} &
\cellcolor{gred!20}\makecell{21\%\\R:61~~B:18\\0.104$\pm$0.031} &
\cellcolor{gred!20}\makecell{14\%\\R:71~~B:15\\0.114$\pm$0.034} &
\cellcolor{gred!20}\makecell{19\%\\R:73~~B:8\\0.112$\pm$0.030} &
\cellcolor{gred!20}\makecell{14\%\\R:72~~B:14\\0.132$\pm$0.053} \\

\midrule
\textbf{0.325} &
\cellcolor{ggreen!20}\makecell{97\%\\R:2~~B:1\\0.049$\pm$0.044} &
\cellcolor{ggreen!20}\makecell{91\%\\R:3~~B:6\\0.060$\pm$0.025} &
\cellcolor{ggreen!20}\makecell{77\%\\R:15~~B:8\\0.066$\pm$0.035} &
\cellcolor{gyellow!20}\makecell{58\%\\R:20~~B:22\\0.085$\pm$0.040} &
\cellcolor{gyellow!20}\makecell{66\%\\R:31~~B:3\\0.082$\pm$0.050} &
\cellcolor{gred!20}\makecell{29\%\\R:59~~B:12\\0.097$\pm$0.038} &
\cellcolor{gred!20}\makecell{17\%\\R:62~~B:21\\0.104$\pm$0.031} \\

\midrule
\textbf{0.350} &
\cellcolor{ggreen!20}\makecell{96\%\\R:0~~B:4\\0.061$\pm$0.068} &
\cellcolor{ggreen!20}\makecell{94\%\\R:4~~B:2\\0.065$\pm$0.088} &
\cellcolor{ggreen!20}\makecell{91\%\\R:3~~B:6\\0.074$\pm$0.098} &
\cellcolor{ggreen!20}\makecell{84\%\\R:8~~B:8\\0.084$\pm$0.095} &
\cellcolor{ggreen!20}\makecell{79\%\\R:13~~B:8\\0.093$\pm$0.107} &
\cellcolor{gyellow!20}\makecell{57\%\\R:25~~B:18\\0.097$\pm$0.059} &
\cellcolor{gred!20}\makecell{28\%\\R:55~~B:17\\0.132$\pm$0.061} \\

\midrule
\textbf{0.375} &
\cellcolor{ggreen!20}\makecell{91\%\\R:4~~B:5\\0.081$\pm$0.038} &
\cellcolor{ggreen!20}\makecell{90\%\\R:4~~B:6\\0.090$\pm$0.137} &
\cellcolor{ggreen!20}\makecell{96\%\\R:1~~B:3\\0.069$\pm$0.048} &
\cellcolor{ggreen!20}\makecell{93\%\\R:1~~B:6\\0.089$\pm$0.126} &
\cellcolor{gyellow!20}\makecell{74\%\\R:7~~B:19\\0.097$\pm$0.067} &
\cellcolor{gyellow!20}\makecell{54\%\\R:16~~B:30\\0.125$\pm$0.107} &
\cellcolor{gred!20}\makecell{17\%\\R:45~~B:37\\0.155$\pm$0.097} \\

\bottomrule
\end{tabular}}}
        \caption{\textbf{Spiral Stairs.}}
        \label{tab:stairs-spiral}
    \end{subtable}
    \caption{\textbf{Traversing Straight and Spiral Stairs.} Each cell reports success rate,
    riser-contact falls (R) and balance falls (B), and mean\,$\pm$\,std 3D
    foot placement error (m). 100 evaluations with initial robot yaw sampled from $\mathcal{U}(-\pi/6,\pi,6)$.}
    \label{tab:stairs-all}
\end{table*}

\textbf{Traversing Cluttered Environments.}~Here the humanoid is tasked with navigating a 6$\times$6~m region with up to 20 traffic cones, as shown in Figure~\ref{fig:tasks}. We employ Anytime Repairing A$^*$ (ARA$^*$) planner~\cite{hornung2012anytime} to generate a sequence of foothold targets. The target candidates are selected from a set defined within the policy's training range. We evaluate scenarios with 10, 15, and 20 traffic cones, conducting 100 trials for each configuration. Table~\ref{tab:clutter_outcomes} shows that our $\pi_{\mathrm{FT}}$ integrates seamlessly with the planner across all tested cases, with only one falling case observed. Failures are primarily due to the planning algorithms not finding a feasible path within the time budget.

\textbf{Traversing Straight Stairs.}~~We evaluate the 3D-trained $\pi_{\mathrm{FT}}$ on a structured obstacle comprising an ascending staircase (5 steps), a landing platform, and a descending staircase (5 steps). $\pi_{\mathrm{FT}}$ was trained with elevation targets sampled in $[-0.15, 0.15]$\,m. We analyze performance over 100 episodes, randomizing the initial yaw orientation in $\mathcal{U}(-\pi/6, \pi/6)$ and varying stair geometry (step length and height). A deterministic planner feeds $\pi_{\mathrm{FT}}$ with 3D footholds matching the terrain profile.
Table~\ref{tab:stairs} shows that as the step height increases, the step length becomes a decisive factor: shorter steps leave little margin for tracking errors, as the swing foot must lift nearly vertically to clear the riser, making even minor inaccuracies likely to cause collisions. Longer steps provide a greater safety margin, allowing the policy to tolerate small placement imperfections without contacting the subsequent riser.
This is further confirmed by the failure mode breakdown. At shorter step lengths, failures are dominated by riser contacts, reflecting the tight clearance constraints. As step length enlarges, riser-contact failures decrease and balance-related falls become the primary failure mode, a consequence of the wider stance required to sustain a longer gait cycle.


\textbf{Traversing Spiral Stairs.}~~We further evaluate the 3D-trained $\pi_{\mathrm{FT}}$ on a more challenging terrain consisting of an ascending staircase with rotating steps ($-10^{\circ}$ yaw offset per step) terminating on a flat platform, designed to emulate a spiral staircase and to validate the omnidirectional capabilities of policies trained within the proposed framework. The policy is identical to that used in the straight staircase experiment. 
Table~\ref{tab:stairs-spiral} shows that the results mirror those of the straight staircase: longer steps consistently reduce riser-contact failures.
However, due to the step rotation, targets are no longer placed at uniform lateral and vertical distances, meaning that further increasing the step length does not monotonically reduce failures. As evidenced by the results, the optimal step length is $0.35$\,m, beyond which balance-related failures begin to increase. Together with the straight staircase experiment, these results confirm that $\pi_{\mathrm{FT}}$ is capable of complex omnidirectional locomotion while operating without terrain perception, relying solely on proprioception.

\begin{table}[t]
\centering
\resizebox{\linewidth}{!}{{\setlength{\tabcolsep}{4pt}
\renewcommand{\arraystretch}{1.3}
\scriptsize
\begin{tabular}{c ccccccc}
\toprule
\multirow{2}{*}{\textbf{\makecell{Step\\Length\\(m)}}}
  & \multicolumn{7}{c}{\textbf{Final Rise (m)}} \\
\cmidrule(lr){2-8}
& \textbf{0.05} & \textbf{0.10} & \textbf{0.15}
& \textbf{0.20} & \textbf{0.25} & \textbf{0.30} & \textbf{0.35} \\
\midrule

\textbf{0.300} &
\cellcolor{ggreen!20}\makecell{100\%\\B:0\\0.039$\pm$0.009} &
\cellcolor{ggreen!20}\makecell{100\%\\B:0\\0.046$\pm$0.007} &
\cellcolor{ggreen!20}\makecell{99\%\\B:1\\0.058$\pm$0.015} &
\cellcolor{ggreen!20}\makecell{76\%\\B:24\\0.133$\pm$0.093} &
\cellcolor{gyellow!20}\makecell{47\%\\B:53\\0.155$\pm$0.085} &
\cellcolor{gred!20}\makecell{7\%\\B:93\\0.319$\pm$0.172} &
\cellcolor{gred!20}\makecell{0\%\\B:100\\0.350$\pm$0.154} \\

\midrule
\textbf{0.325} &
\cellcolor{ggreen!20}\makecell{100\%\\B:0\\0.037$\pm$0.004} &
\cellcolor{ggreen!20}\makecell{100\%\\B:0\\0.043$\pm$0.009} &
\cellcolor{ggreen!20}\makecell{99\%\\B:1\\0.057$\pm$0.015} &
\cellcolor{ggreen!20}\makecell{96\%\\B:4\\0.092$\pm$0.071} &
\cellcolor{ggreen!20}\makecell{87\%\\B:13\\0.120$\pm$0.060} &
\cellcolor{gyellow!20}\makecell{35\%\\B:65\\0.230$\pm$0.140} &
\cellcolor{gred!20}\makecell{5\%\\B:95\\0.338$\pm$0.151} \\

\midrule
\textbf{0.350} &
\cellcolor{ggreen!20}\makecell{100\%\\B:0\\0.040$\pm$0.011} &
\cellcolor{ggreen!20}\makecell{100\%\\B:0\\0.045$\pm$0.010} &
\cellcolor{ggreen!20}\makecell{97\%\\B:3\\0.064$\pm$0.049} &
\cellcolor{ggreen!20}\makecell{96\%\\B:4\\0.090$\pm$0.051} &
\cellcolor{ggreen!20}\makecell{89\%\\B:11\\0.110$\pm$0.057} &
\cellcolor{gyellow!20}\makecell{68\%\\B:32\\0.199$\pm$0.113} &
\cellcolor{gred!20}\makecell{6\%\\B:94\\0.380$\pm$0.170} \\

\midrule
\textbf{0.375} &
\cellcolor{ggreen!20}\makecell{100\%\\B:0\\0.054$\pm$0.019} &
\cellcolor{ggreen!20}\makecell{100\%\\B:0\\0.065$\pm$0.016} &
\cellcolor{ggreen!20}\makecell{95\%\\B:5\\0.104$\pm$0.060} &
\cellcolor{ggreen!20}\makecell{85\%\\B:15\\0.115$\pm$0.064} &
\cellcolor{gyellow!20}\makecell{47\%\\B:53\\0.173$\pm$0.065} &
\cellcolor{gyellow!20}\makecell{46\%\\B:54\\0.207$\pm$0.101} &
\cellcolor{gyellow!20}\makecell{57\%\\B:43\\0.201$\pm$0.103} \\

\bottomrule
\end{tabular}}}
\caption{\textbf{Traversing Ramps.} Each cell reports success rate, balance falls (B), and mean\,$\pm$\,std 3D foot placement error (m). 100 evaluations with initial robot yaw $\sim\mathcal{U}(-\pi/6,\pi/6)$).}
\label{tab:ramp}
\end{table}

\textbf{Traversing Ramps.}~~Finally, we evaluate $\pi_{\mathrm{FT}}$ on an additional obstacle, structurally analogous to the straight staircase scenario, but replacing discrete steps with continuous ascending and descending ramps of varying inclination. 
This experiment is critical for assessing the generalization capabilities of a policy trained within the proposed framework. Indeed, as described in Section~\ref{sec:method}, the 3D fine-tuning stage employs moving pillars as contact surfaces, meaning the policy never encounters sloped foot placements during training. Successfully traversing ramps therefore requires the policy to generalize its $z$-tracking capabilities to unseen foot pitch configurations.
Table~\ref{tab:ramp} shows that $\pi_{\mathrm{FT}}$ handles this setting effectively. A trend consistent with the staircase experiments emerges: longer steps yield higher success rates as task difficulty increases. However, this phenomenon has a different explanation here. At high inclinations with short steps, the robot loses balance due to the steep vertical rise demanded per foothold. Conversely, excessively long steps also increase balance failures, as the robot must sustain a longer gait cycle. The optimal step length sits around $0.35$\,m across most inclinations. These results confirm that the proposed framework produces policies with strong generalization capabilities.
\begin{figure*}[t]
    \centering
    \resizebox{.62\linewidth}{!}{\includegraphics[]{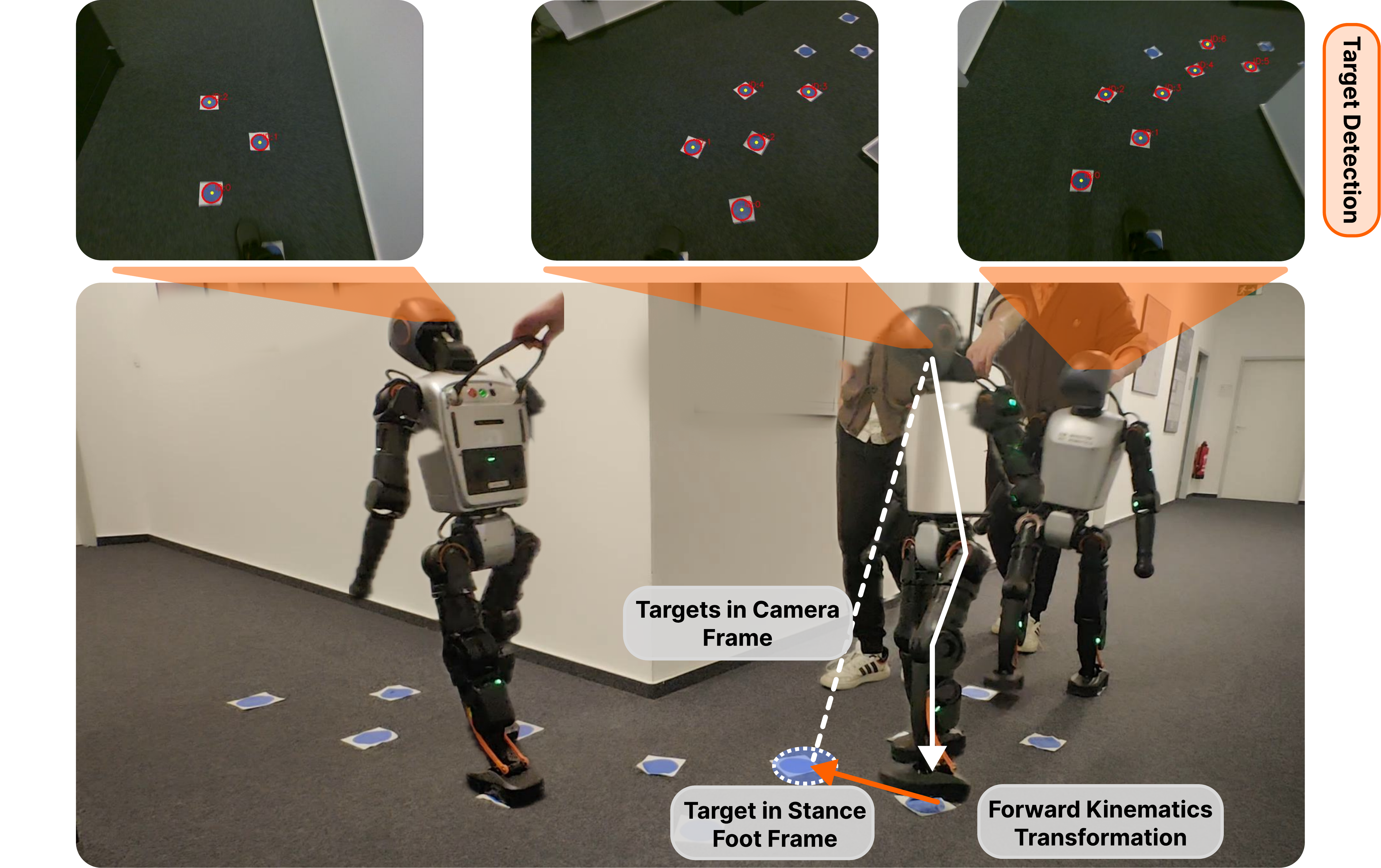}}
    \caption{\textbf{Real-World Deployment.} Sequence of the robot tasked with traversing a corridor while stepping on predefined targets. Markers are detected via the onboard camera and then mapped to valid footholds targets in the stance foot frame via a transformation by forward kinematics. For each pose snapshot, we display the corresponding camera frame with highlighted targets. The second pose from the right details the target generation logic during a gait phase switch. Here, the right foot becomes the new stance foot.}
    \label{fig:real-w}
    \vspace{-1em}
\end{figure*} 

\section{Real-World Results} \label{sec:exp-real}

In the real-world experiment, to avoid additional complexity introduced by the localization problem, we employ a vision-based foothold tracking task to verify our FT policy. As illustrated in Figure~\ref{fig:real-w}, a sequence of blue circular markers is manually placed on the ground in a corner of a corridor to serve as visual foothold targets. We use the onboard RealSense D455 camera of the Booster T1 robot to extract the targets. By fusing RGB-based target detection, the depth measurements, and the robot's kinematic model, we compute the relative transformations between the targets and the robot's current stance foot. The foothold orientations are heuristically determined based on the upcoming swing targets to ensure the feasibility of the future targets. A snapshot of the target detection results is illustrated in Figure~\ref{fig:real-w}. 
Due to delays introduced by camera frame rate limitations, communication latency, and policy inference time, the swing foot may not reach the commanded foothold exactly at the nominal swing–stance transition defined by the gait phase. To address this issue, we implement a simple delay compensation mechanism that repeats the same gait phase $\bm{\phi}_t$ and foothold target $\bm{g}_t$ until the relative height between the swing and the stance foot falls below a threshold, ensuring the relative targets are computed with the next stance on the ground. 

\textbf{Results}. We evaluate the system over 20 real-world trials, including both forward and backward traversals in the corridor with manual resets. An uncut video of the experiments is provided as supplementary. Despite sensing noise and system delays, the robot achieves 93.08\% success of stepping on designated foothold targets while maintaining stable locomotion. These results demonstrate the practical viability of the proposed FT controller in real-world settings, even when driven by imperfect visual perception. The primary sources of foot-placement error arise from motion-induced image blur, which degrades target detection accuracy, and from depth sensing artifacts, as the depth camera is not optimized for highly dynamic motion.  Furthermore, although the sim-to-real gap is significantly mitigated by the large-scale domain randomization, it also poses costs in losing tracking accuracy. 
Despite the presence of tracking error, 
this experiment successfully verifies that $\pi_{\mathrm{FT}}$ can be seamlessly integrated into sophisticated high-level planners to support various downstream tasks.
\section{Conclusion} \label{sec:conclusion}
In this work, we present a general foothold-conditioned training framework that integrates seamlessly with a wide range of high-level planning algorithms. By introducing a general-purpose 3D goal sampler, our approach enables adaptive terrain generation and supports stable foot placement. To address the limitations of prior methods that rely on accurate pose estimation or explicit foot–contact estimation, we represent foothold targets in the stance-foot frame and hold them constant throughout each swing phase. This design ensures consistent target tracking and robust execution. 
Through extensive comparisons with a standard velocity-tracking policy, we show that our foothold tracking controller achieves more agile and efficient goal reaching, delivers consistent performance in precision-critical tasks, and robustly traverses challenging cluttered environments and staircases. Finally, we validate the real-world deployability of our approach through a fully vision-based foothold tracking experiment, demonstrating reliable sim-to-real transfer without reliance on external localization. Together, these results highlight the practicality, versatility, and robustness of the proposed framework for humanoid locomotion in complex safety-critical scenarios.
\clearpage

\setlength{\textfloatsep}{16pt}

\section*{Acknowledgments}
This publication was funded with the contribution of Ministero dell’Università e della ricerca pursuant to D.D. n. 7206 of 17 April 2025 – BANDO FIS 2. Project FIS-2023-02598 (Starting Grant), title: “Unified Learning from Diverse Human Feedback” (HUmLrn). CUP: D53C25000710001. The project is supported by the Fundamental Research Funds for the Central Universities. 



\bibliographystyle{plainnat}
\bibliography{references}


\clearpage
\onecolumn

\appendix

Here, we present additional details for the experiments presented in the main paper.
All the media reporting videos for the real-world experiment and the simulated ones can be found in the attached zip and in the project website (\href{https://montenegroalessandro.github.io/mind-your-steps/}{\textcolor{gblue}{\textbf{\texttt{https://montenegroalessandro.github.io/mind-your-steps/}}}}).

\subsection{Observations for Actor and Critic Networks}
The table below details the observations received by the actor and critic networks, specifying whether the components are subject to noise.

\begin{center}
    \centering
    \resizebox{.33\linewidth}{!}{\begin{tabular}{llc}
    \toprule
    \textbf{Network} & \textbf{Observation} & \textbf{Noise} \\
    \midrule
    \multirow{5}{*}{\textbf{Actor}}
    & Base Angular Velocity & \cmark \\
    & Projected Gravity & \cmark \\
    & Joint Positions & \cmark \\
    & Joint Velocities & \cmark \\
    & Last Action & \xmark \\
    & Goal & \xmark \\
    \midrule
    \multirow{5}{*}{\textbf{Critic}} 
    & Base Linear Velocity & \xmark \\
    & Base Angular Velocity & \xmark \\
    & Projected Gravity & \xmark \\
    & Joint Positions & \xmark \\
    & Joint Velocities & \xmark \\
    & Last Action & \xmark \\
    & Goal & \xmark \\
    \bottomrule
    \end{tabular}}
\end{center}

\subsection{Domain Randomization}
We detail the domain randomization parameters below. Randomization is applied at the beginning of each episode, with the exception of external disturbances, which occur at each timestep with a certain probability.
For each parameter, we define the randomization method as follows:
\begin{itemize}
    \item \textbf{Direct}: The value is uniformly sampled from the specified interval and directly employed.
    \item \textbf{Multiplier}: A scaling factor is uniformly sampled from the reported range and multiplied by the nominal value.
    \item \textbf{Additive $\mathcal{U}$}: An offset is sampled uniformly from the range and added to the nominal value.
    \item \textbf{Additive $\mathcal{N}$}: Noise is sampled from a Normal distribution with the specified scale and added to the nominal value.
\end{itemize}

\begin{center}
    \resizebox{.55\linewidth}{!}{\begin{tabular}{lcc}
    \toprule
    \textbf{Parameter} & \textbf{Type} & \textbf{Value / Range} \\
    \midrule
    \multicolumn{3}{c}{\textit{Global Dynamics \& Inertial Properties}} \\
    Gravity ($z$-axis) & Direct & $[9.51, 10.11] \, \text{m/s}^2$ \\
    Root Body Mass & Multiplier & $[0.8, 1.2]$ \\
    Link Mass & Multiplier & $[0.9, 1.1]$ \\
    CoM Displacement ($x, y, z$) & Additive $\mathcal{U}$ & $[-0.05, 0.05] \, \text{m}$ \\
    \midrule
    \multicolumn{3}{c}{\textit{Terrain Contact Properties (Floor \& Pillars)}} \\
    Tangential Friction & Direct & $[0.5, 1.5]$ \\
    Torsional Friction & Direct & $[0.1, 0.3]$ \\
    Rolling Friction & Direct & $[8 \times 10^{-5}, 1.2 \times 10^{-4}]$ \\
    \midrule
    \multicolumn{3}{c}{\textit{Actuation \& Joint Properties}} \\
    Joint Damping & Direct & $[0.005, 0.03] \, \text{Nms/rad}$ \\
    Joint Friction Loss & Direct & $[0.0, 1.0] \, \text{Nm}$ \\
    Joint Armature & Direct & $[0.007, 0.03] \, \text{kg m}^2$ \\
    KP Gain Noise & Additive $\mathcal{U}$ & $[-0.15 \cdot K_{P}, 0.15 \cdot K_{P}]$ \\
    KD Gain Noise & Additive $\mathcal{U}$ & $[-0.5 \cdot K_{D}, 0.5 \cdot K_{D}]$\\
    \midrule
    \multicolumn{3}{c}{\textit{Observation Noise}} \\
    Joint Position & Additive $\mathcal{N}$ & $\pm 0.03 \, \text{rad}$\\
    Joint Velocity & Additive $\mathcal{N}$ & $\pm 0.3 \, \text{rad/s}$ \\
    Angular Velocity & Additive $\mathcal{N}$ & $\pm 0.2 \, \text{rad/s}$\\
    Projected Gravity Vector & Additive $\mathcal{N}$ & $\pm 0.015$ \\
    \midrule
    \multicolumn{3}{c}{\textit{External Disturbances}} \\
    Random Kicks (Velocity) & \makecell{Additive $\mathcal{U}$\\{\small (Per-step w.p. $0.004$)}} & $[0.1, 0.4] \, \text{m/s}$ \\
    \bottomrule
    \end{tabular}}
\end{center}

\subsection{PPO Common Training Hyperparameters}
The table below details the PPO hyperparameters common to both the flat-terrain and $z$-tracking training configurations.

\begin{center}
    \centering
    \resizebox{.35\linewidth}{!}{\begin{tabular}{lc}
        \toprule
        \textbf{Parameter} & \textbf{Value} \\
        \midrule
        \multicolumn{2}{c}{\textit{PPO Algorithm}} \\
        Discount Factor ($\gamma$) & $0.995$ \\
        GAE Parameter ($\lambda$) & $0.95$ \\
        Clipping Range ($\epsilon$) & $0.2$ \\
        Entropy Coefficient & $0.01$ \\
        Value Function Coefficient & $0.5$ \\
        Max Gradient Norm & $1.0$ \\
        \midrule
        \multicolumn{2}{c}{\textit{Adaptive Learning Rate}} \\
        Initial Learning Rate & $1 \times 10^{-5}$ \\
        Min Learning Rate & $1 \times 10^{-6}$ \\
        Max Learning Rate & $0.01$ \\
        KL Target & $0.02$ \\
        KL Margin & $1.5$ \\
        KL Scale Factor & $1.5$ \\
        \midrule
        \multicolumn{2}{c}{\textit{Employed Samples}} \\
        Parallel Environments & $8192$\\
        Horizon Steps & $50$ \\
        Update Epochs & $20$ \\
        Mini-batches per Epoch & $1$ \\
        Observation Normalization & True \\
        Reward Normalization & True \\
        Observation History Length & $1$ \\
        \midrule
        \multicolumn{2}{c}{\textit{Policy Network}} \\
        Hidden Layers & $512 \times 256 \times 128$ \\
        Activation Function & ELU \\
        Initial Action Std & $0.135$ \\
        Learnable Action Std & True \\
        \bottomrule
    \end{tabular}}
\end{center}

\subsection{Additional Reward Terms} \label{apx:rew}
While Section~\ref{sec:method} details the main tracking terms, \ie $r_{\text{track}}$ and $r_{\text{feet}}$, here we specify the additional reward terms employed and report the common used coefficients' values (Table~\ref{tab:reward-terms}). For other penalization terms, please refer to the \texttt{CrispBoosterLocomotionReward} in~\citep{alhafez2023locomujoco}, upon which our implementation is based.


\begin{center}
    \centering
    \renewcommand{\arraystretch}{1.6}
    \setlength{\tabcolsep}{8pt}
    \resizebox{.8\linewidth}{!}{
    \begin{tabular}{l l c}
        \toprule
        \textbf{Reward Term} & \textbf{Description} & \textbf{Coefficients} \\
        \midrule
        
        $r_{\text{jnt}}(\bm{s}_{t},\bm{a}_{t}) = \omega_{\text{jnt}} \exp\left( -\xi_{\text{jnt}} \| \bm{q}_{\text{jnt},t} - \bar{\bm{q}}_{\text{jnt}} \|_{2}^{2} \right)$
        & Tracks upper body joint reference $\bar{\bm{q}}_{\text{jnt}}$.
        & $\omega_{\text{jnt}}= \xi_{\text{jnt}}= 4$ \\
        
        $r_{\text{bh}}(\bm{s}_{t},\bm{a}_{t}) = -\omega_{\text{bh}} \left(\prescript{w}{b}{\bm{p}}_{t,z} - \prescript{w}{b}{\bar{\bm{p}}}_{z} \right)^{2}$
        & Tracks CoM height reference $\prescript{w}{b}{\bar{\bm{p}}}_{z}$.
        & $\omega_{\text{bh}}=10$ \\
        
        $r_{\text{ar}}(\bm{s}_{t}, \bm{a}_{t}) = -\omega_{\text{ar}} \| \bm{a}_{t} - \bm{a}_{t-1} \|_{2}^{2}$
        & Penalizes large action changes (smoothness).
        & $\omega_{\text{ar}} = 3$ \\
        
        $r_{\text{fs}}(\bm{s}_{t}, \bm{a}_{t}) = -\omega_{\text{fs}} \sum_{\Box \in \{l,r\}} \| \prescript{w}{\Box}{\dot{\bm{p}}}_{t} \cdot \mathds{C}_{\Box} \|_{2}^{2}$
        & Penalizes foot velocity during ground contact (no-slip).
        & $\omega_{\text{fs}} = 4$ \\
        
        \bottomrule
    \end{tabular}}
    \captionof{table}{Additional reward terms and their coefficients.}
    \label{tab:reward-terms}
\end{center}

For the sake of completeness, we additionally report (Table~\ref{tab:reward-tracking}) the terms we detailed in Section~\ref{sec:method}.


\begin{center}
    \centering
    \renewcommand{\arraystretch}{1.6}
    \setlength{\tabcolsep}{8pt}
    \resizebox{.8\linewidth}{!}{
    \begin{tabular}{l l c}
        \toprule
        \textbf{Reward Term} & \textbf{Description} & \textbf{Coefficients} \\
        \midrule
        
        \multirow{1}{*}{$r_{\text{feet}}(\bm{s}_{t}, \bm{a}_{t})$ (See Section~\ref{sec:method})}
        & \multirow{1}{*}{Enforces feet swinging in the allotted period.}
        & $\omega_{\text{feet}} = 6$, $\delta_{\text{sw}} = 0.1$ \\
        
        \midrule
        
        \multirow{2}{*}{$r_{\text{track}}(\bm{s}_{t}, \bm{a}_{t})$ (See Section~\ref{sec:method})}
        & \multirow{2}{*}{Tracks desired foothold targets.}
        & $\omega_{1} = \omega_{2} = \omega_{3} = 5$ \\
        & & $\xi_{1} = \xi_{3} = 100$, $\xi_{2} = 200$ \\

        \midrule
        
        \multirow{2}{*}{$r_{\text{k}}(\bm{s}_{t}, \bm{a}_{t})$ (See Section~\ref{sec:method})}
        & \multirow{2}{*}{Enforces the knee of the swinging foot to stay above a certain height.}
        & $\omega_{\text{k}} = 4$, $\xi_{k} = 200$ \\
        & & $\delta_{k} = 0.25$ \\
        
        \bottomrule
    \end{tabular}}
    \captionof{table}{Tracking and feet swing reward terms and their coefficients.}
    \label{tab:reward-tracking}
\end{center}

\subsection{Details on the Employed Velocity-based Policy}
In Section~\ref{sec:exp-sim}, we compare our approach against a velocity-tracking policy $\pi_{\text{VT}}$, trained via the \emph{LocoMuJoCo} pipeline~\citep{alhafez2023locomujoco}. Specifically, $\pi_{\text{VT}}$ shares the same observation vector as our $\pi_{\text{FT}}$:
\begin{align*}
    \bm{x}_{t} = \left( \bm{o}_{\text{proprio},t}; \bm{a}_{t-1}; \bm{\phi}_{t}; \bm{g}_{t} \right),
\end{align*}
where the goal $\bm{g}_{t}$ specifies the target base velocity and height:
\begin{align*}
    \bm{g}_{t} = \left( \prescript{w}{b}{\dot{\bar{\bm{p}}}}_{x}; \prescript{w}{b}{\dot{\bar{\bm{p}}}}_{y}; \prescript{w}{b}{\dot{\bar{\psi}}}; \prescript{w}{b}{\bar{\bm{p}}}_{z} \right).
\end{align*}
Here, $\prescript{w}{b}{\dot{\bar{\bm{p}}}}_{x}$ and $\prescript{w}{b}{\dot{\bar{\bm{p}}}}_{y}$ denote the target longitudinal and lateral velocities, $\prescript{w}{b}{\dot{\bar{\psi}}}$ is the target angular velocity, and $\prescript{w}{b}{\bar{\bm{p}}}_{z}$ is the target base height.
We highlight an additional difference regarding the primary tracking objectives compared to $\pi_{\text{FT}}$. While the foothold-tracking policy aims at minimizing the error terms detailed in Section~\ref{sec:method}, $\pi_{\text{VT}}$ utilizes a standard velocity-tracking reward defined as:
\begin{align*}
    r_{\text{track}}(\bm{s}_{t}, \bm{a}_{t}) &= \omega_{1} \exp \left( -\xi_{1} (\prescript{w}{b}{\dot{\bm{p}}}_{t,x} - \prescript{w}{b}{\dot{\bar{\bm{p}}}}_{x})^{2} \right) \\
    &\quad + \omega_{2} \exp \left( -\xi_{2} (\prescript{w}{b}{\dot{\bm{p}}}_{t,y} - \prescript{w}{b}{\dot{\bar{\bm{p}}}}_{y})^{2} \right) \\
    &\quad + \omega_{3} \exp \left( -\xi_{3} (\prescript{w}{b}{\dot{\psi}}_{t} - \prescript{w}{b}{\dot{\bar{\psi}}})^{2} \right),
\end{align*}
with coefficients set to $\omega_{1} = \omega_{2} = 1.5$, $\omega_{3} = 1$, and $\xi_{1} = \xi_{2} = \xi_{3} = 4$.
Base height tracking follows the formulation described in Appendix~\ref{apx:rew}, while all other regularization terms remain identical to those used for $\pi_{\text{FT}}$. Finally, the target velocities are sampled uniformly: $\prescript{w}{b}{\dot{\bar{\bm{p}}}}_{x} \sim \mathcal{U}(-0.8, 0.8)$ m/s, $\prescript{w}{b}{\dot{\bar{\bm{p}}}}_{y} \sim \mathcal{U}(-0.5, 0.5)$ m/s, and $\prescript{w}{b}{\dot{\bar{\psi}}} \sim \mathcal{U}(-1.5, 1.5)$ rad/s.

\subsection{Training Parameters for the Policy Used for Flat Terrains}
The following details hold for both $\pi_{\mathrm{FT}}$ and $\pi_{\mathrm{FT-R}}$, i.e., the FT policy with targets represented int he stance foot frame and in the root frame, respectively. The policies were trained with PPO for a total of $10^{9}$ timesteps. The table below reports the values used for the goal parameters. Parameters defined by a range are sampled uniformly from the specified interval according to the schedule described in Section~\ref{sec:method}, while scalar values denote parameters kept constant throughout training.

\begin{center}
    \resizebox{.35\linewidth}{!}{\begin{tabular}{lc}
    \toprule
    \textbf{Parameter} & \textbf{Value / Range} \\
    \midrule
    \multicolumn{2}{c}{\textit{Step Generation}} \\
    Movement Direction ($\theta_{\text{move}}$) & $[-\pi, \pi)$ rad \\
    Feet Direction ($\theta_{\text{feet}}$) & $0$ rad \\
    Step Length ($d$) & $[0.2, 0.5]$ m \\
    $\theta_{\text{move}}$ Perturbation ($\alpha$) & $[-2 \pi / 9, 2 \pi / 9]$ rad \\
    $\theta_{\text{feet}}$ Perturbation ($\beta$) & $[-\pi/6, \pi/6]$ rad \\
    Height Offset ($z$) & $0$ m \\
    \midrule
    \multicolumn{2}{c}{\textit{Gait Constraints}} \\
    Minimum Feet Distance & $0.10$ m \\
    Hold Still Probability ($p_{\text{hold}}$) & $0.1$ \\
    Hold Still Feet Width & $0.20$ m \\
    \bottomrule
    \end{tabular}}
\end{center}

For the reward terms, we employed the values detailed in Appendix~\ref{apx:rew}, but setting $\omega_{2} = 0$ in order to exclude the tracking of the $z$-component. Moreover, we set $\omega_{\text{k}}=0$ in order to exclude the swinging knee height term.

\subsection{Training Parameters for the Policy Used for Stairs}

In this experiment, the policy was trained using PPO for $1.2 \cdot 10^{9}$ timesteps, initialized from the flat-terrain policy weights. The table below lists the goal parameters. Ranges imply uniform sampling according to the schedule in Section~\ref{sec:method}, while scalar values denote constant parameters. The reward terms are detailed in Appendix~\ref{apx:rew}.

\begin{center}
    \resizebox{.35\linewidth}{!}{\begin{tabular}{lc}
    \toprule
    \textbf{Parameter} & \textbf{Value / Range} \\
    \midrule
    \multicolumn{2}{c}{\textit{Step Generation}} \\
    Movement Direction ($\theta_{\text{move}}$) & $\pi x$ rad, $x \sim \mathcal{B}(0.5)$ \\
    Feet Direction ($\theta_{\text{feet}}$) & $0$ rad \\
    Step Length ($d$) & $[0.2, 0.4]$ m \\
    $\theta_{\text{move}}$ Perturbation ($\alpha$) & $[-\pi / 18, \pi / 18]$ rad \\
    $\theta_{\text{feet}}$ Perturbation ($\beta$) & $[-\pi/18, \pi/18]$ rad \\
    Height Offset ($z$) & $[-0.15, 0.15]$ m \\
    \midrule
    \multicolumn{2}{c}{\textit{Gait Constraints}} \\
    Minimum Feet Distance & $0.10$ m \\
    Hold Still Probability ($p_{\text{hold}}$) & $0$ \\
    Hold Still Feet Width & $0.20$ m \\
    \bottomrule
    \end{tabular}}
\end{center}

Notably, for this task, we restricted the motion to strictly forward or backward directions, without the possibility of staying still. This is denoted in the table by $\theta_{\text{move}} = \pi x$, where $x \sim \mathcal{B}(0.5)$ represents a Bernoulli variable yielding discrete values $\{0, 1\}$ with equal probability. Notice that the fact that the robot is just trained to go forward or backward it is not restricting the robot's curving capabilities, as demonstrated by the spiral stairs experiment.

\textbf{On Moving Pillars Terrain.}~~To ensure continuous foothold availability, we employed a dynamic terrain system comprising three dynamic pillars, with a diameter of $0.3$ m and friction features matching the floor. These pillars are placed such that whenever a new foot placement target is sampled, the currently free pillar is relocated to the goal position. We define a pillar as free when it is not currently supporting a foot. To ensure valid placements, we employ a rejection sampling mechanism ($2048$ batched samples) that prevents collisions with the robot's feet or the active stance pillars. Furthermore, to mitigate early termination caused by minor planar tracking errors, we implement an adaptive landing mechanism. During the final portion of the swing phase (specifically when $\phi_{t} \in [0.25 + \delta_{\text{sw}}, 0.5)$ or $\phi_{t} \in [0.75 + \delta_{\text{sw}}, 1]$, see Figure~\ref{fig:swing}), the target pillar shifts its $xy$-position to track the swinging foot's projection. Crucially, the pillar's height $z$ remains fixed to the originally sampled value. 

\subsection{High-Level Planners}
In this section, we detail the high-level planners employed for both the simulated experiments and the real-world deployment.

\textbf{Goal-Reaching Planner.}~~For the goal-reaching task, $\pi_{\text{FT}}$ is guided by a deterministic foothold planner. At each gait update, the goal position and base position are expressed in the \emph{stance-foot frame} as $\prescript{s}{g}{\bm{p}}$ and $\prescript{s}{b}{\bm{p}}$, respectively.
If the planar Euclidean distance to the goal falls below a threshold  $\| \prescript{s}{g}{\bm{p}}_{xy} - \prescript{s}{b}{\bm{p}}_{xy} \|_{2} \leq \epsilon$, the robot is commanded to stand still, marking task completion.
Otherwise, the target for the next swing foot is generated based on the difference vector $\bm{d} = \prescript{s}{g}{\bm{p}} - \prescript{s}{b}{\bm{p}}$. The target yaw is computed as $\theta=\text{clip}(\text{atan2}(\bm{d}_{y},\bm{d}_{x}),-60^\circ,60^\circ)$, while the foot target is calculated as $^s_\square\bm{p} = \frac{1}{2} \mathbb{I}_{\text{s}} d_{\text{feet}}\bm{v}_{\text{feet}} +  d_{\text{max}} \frac{\bm{d}}{\|\bm{d}\|_{2}}$, where $d_{\text{feet}}$ prescribes the nominal distance between two feet, $d_{\text{max}}$ is the maximum distance per step, $\bm{v}_{\text{feet}} = [0, 1]^\intercal$, and $\mathbb{I}_{\text{s}} \in \{-1,1\}$ is indicator of the swing foot, $-1$ for right foot swinging and $1$ for left foot swinging.
Notice that for $\pi_{\mathrm{FT-R}}$ the same procedure applies, but then the policy is fed with the goal $\bm{g}_{t}$ components expressed in the \emph{root frame}.
The resulting swing target is clipped to satisfy the feasible constraints to avoid leg crossing. Finally, the planner constructs the goal vector $\bm{g}_t$ in the stance-foot frame as described in Section~\ref{sec:method}: the stance foot reference is set to zero position offset and identity orientation quaternion, while the swing foot reference is the one computed by the planner.
Conversely, for the velocity-tracking baseline $\pi_{\text{VT}}$, a P-controller converts the body-frame planar position error into linear velocity commands (scaled to respect limits), while a separate clipped P-controller generates the yaw rate $\omega_z$ to align the heading with the goal. The maximum velocity of the VT baseline is clamped to the trained range comparable to the FT policy. The maximum velocity $v_{\max}$ is computed by the maximum step distance $d_{\max}$ and gait frequency $f_{\text{gait}}$, i.e., $v_{\max} = d_{\max}/f_{\text{gait}}$. We swept the P-gain to choose the best performing one. 

\textbf{Narrow Bridge Traversal.}~~This task comprises an initial platform, a central narrow bridge, and a final goal platform. While the \emph{approach} and \emph{run-out} sections utilize the previously described goal-reaching planner, the bridge traversal requires a modified planner for lateral walking.
The path is defined by a set of ordered waypoints $\mathcal{P} = \{\bm{p}_{i}\}_{i=0}^{N-1}$, tracked via a progress index $i$.
To traverse the bridge, the agent adopts a lateral gait strategy. Consequently, the planner sets the target yaw perpendicular to the local path segment $(\bm{p}_{i-1}\!\rightarrow\!\bm{p}_i)$, subject to clipping for meeting yaw constraints. Simultaneously, the foothold position is projected onto the segment connecting $\bm{p}_{i-1}$ and $\bm{p}_i$, ensuring strict alignment with the bridge centerline.

\textbf{Cluttered Environment Traversal.}~~In this task, the robot is task with reaching a goal position while avoiding obstacles populating the navigation environment. To generate collision-free foothold targets, we employ a search-based footstep planner~\citep{hornung2012anytime}.
The planner operates on a discretized lattice of stance-foot poses $\bm{s}=(x,y,\theta)$, expanding states via alternating support and relative step actions $\bm{a}=(\Delta x,\Delta y,\Delta\theta)$.
The transition model utilizes a finite action set $\mathcal{A}$ defined by discrete parameters: $\Delta x \in \{0.0,\,0.10,\,0.20\}$ m, $\Delta y \in \{0.20,\,0.25,\,0.30,\,0.35\}$ m, and $\Delta\theta \in \{0,\,\pm10^\circ,\,\pm20^\circ,\,\pm30^\circ\}$.
Given a stance pose $\bm{s}$ and an action $\bm{a} \in \mathcal{A}$, a successor state $\bm{s}' = t(\bm{s},\bm{a})$ is generated and snapped to the lattice resolution for hashing.
Candidate transitions are filtered by feasibility constraints, including step-length bounds and a minimum inter-foot lateral spacing under alternating support. 
Furthermore, collision checking discards successors where the foot geometry overlaps with environmental obstacles.
Node expansion is prioritized using a Euclidean distance heuristic $h(\bm{s})=\| \bm{p}_b - \bm{p}_g \|_{2}$, estimating the cost to the goal $\bm{p}_g$.
The search algorithm minimizes a weighted evaluation function $f(\bm{s})=g(\bm{s})+w\,h(\bm{s})$, where $g(\bm{s})$ represents the accumulated path cost from the start node (with weight $w=5$).
Finally, the computed foothold sequence is encoded into the goal observation $\bm{g}_{t}$ for the policy, as detailed in Section~\ref{sec:method}.

\textbf{Stairs Traversal.}~~To evaluate the policy on the stairs climbing task, we implemented a deterministic high-level planner that supplies foothold targets. 
The obstacle geometry is parameterized by the \textit{step length}, defining the stairs' steps depth; the \textit{step height}, defining the stairs' steps height; the \textit{platform length}, specifying the flat landing separating the ascending and descending sections; and the \textit{feet width}, determining the lateral spacing between the left and right foot trajectories. Additionally, we define a set of \textit{approaching steps} on flat terrain before and after the staircase to stabilize the gait.
Based on these dimensions, the planner pre-computes two distinct, parallel sequences of 3D foothold coordinates, $\mathcal{P}_{l} = \{\prescript{w}{l}{\bm{p}_{i}}\}_{i=0}^{N-1}$ and $\mathcal{P}_{r} = \{\prescript{w}{r}{\bm{p}_{i}}\}_{i=0}^{N-1}$, for the left and right feet respectively. The sequence follows the fixed topology enforced by the obstacle's terrain profile: an initial flat approach, an ascending staircase, a central landing platform, a descending staircase, and a final flat run-out.
During execution, the robot tracks its progress using two indices, $i_l$ and $i_r$, pointing to the current target in $\mathcal{P}_{l}$ and $\mathcal{P}_{r}$. The starting swing foot is selected based on the robot's initial yaw orientation. 
At each gait phase switch (swing $\to$ stance), the indices are updated to enforce an alternating gait: if the right foot is becoming the swing foot, its target index advances as $i_r \leftarrow i_l + 1$; conversely, if the left foot swings, $i_l \leftarrow i_r + 1$.
Finally, the planner outputs the goal $\bm{g}_t$ in the \textit{stance-foot frame}, as described in Section~\ref{sec:method}. Specifically, the reference for the incoming stance foot is set to a zero position and identity orientation offset, while the reference for the incoming swing foot is computed as the relative transformation between the current stance foot and the target foothold retrieved from the plan.

\textbf{Real-World Vision-Based Planner.}~~As mentioned in Section~\ref{sec:exp-real}, the hardware experiments on the Booster T1 humanoid rely on a high-level planner driven by a computer vision module exploiting the onboard camera.
The task involves navigating a corridor by stepping onto manually placed circular markers.
We employ the onboard color camera, a RealSense D455, to detect these targets. Specifically, at each gait switch instant, the planner fuses RGB detection with depth measurements to extract the next foothold position in the camera frame.
Using the robot's kinematics, this target is transformed into the current stance-foot frame to compute the relative position offsets required by the policy as detailed in Section~\ref{sec:method} (keeping the stance foot offsets at zero). Orientation targets are determined heuristically to ensure kinematic feasibility.
Finally, as already described in Section~\ref{sec:exp-real}, to account for real-world latencies and imperfect ground contacts, we do not adhere strictly to the open-loop gait clock. Instead, the gait signal is frozen (repeating the last value) until the relative vertical distance between the stance and swing feet falls below a safety threshold, ensuring robust ground contact before transitioning.

\subsection{Section~\ref{sec:exp-sim}'s Extended Tables}
Here, we report extended tables for the experiments conducted in Section~\ref{sec:exp-sim}. Specifically, we include results for the straight and spiral staircase traversal experiments.



    
    

\begin{table*}[h]
    \centering
    \resizebox{.8\linewidth}{!}{\setlength{\tabcolsep}{3pt}
\renewcommand{\arraystretch}{1.3}
\tiny
\begin{tabular}{c cccccccccc}
\toprule
\multirow{2}{*}{\textbf{\makecell{Step\\Length\\(m)}}}
  & \multicolumn{10}{c}{\textbf{Step Height (m)}} \\
\cmidrule(lr){2-11}
& \textbf{0.01} & \textbf{0.02} & \textbf{0.04} & \textbf{0.08}
& \textbf{0.10} & \textbf{0.11} & \textbf{0.115}
& \textbf{0.12} & \textbf{0.125} & \textbf{0.13} \\
\midrule

\textbf{0.300} &
\cellcolor{ggreen!20}\makecell{100\%\\R:0~~B:0\\0.033$\pm$0.006} &
\cellcolor{ggreen!20}\makecell{100\%\\R:0~~B:0\\0.034$\pm$0.003} &
\cellcolor{ggreen!20}\makecell{100\%\\R:0~~B:0\\0.036$\pm$0.005} &
\cellcolor{ggreen!20}\makecell{95\%\\R:3~~B:2\\0.059$\pm$0.089} &
\cellcolor{ggreen!20}\makecell{76\%\\R:19~~B:5\\0.071$\pm$0.049} &
\cellcolor{gyellow!20}\makecell{60\%\\R:40~~B:0\\0.081$\pm$0.048} &
\cellcolor{gyellow!20}\makecell{52\%\\R:34~~B:14\\0.090$\pm$0.048} &
\cellcolor{gyellow!20}\makecell{56\%\\R:39~~B:5\\0.083$\pm$0.041} &
\cellcolor{gred!20}\makecell{38\%\\R:47~~B:15\\0.095$\pm$0.038} &
\cellcolor{gred!20}\makecell{12\%\\R:70~~B:18\\0.108$\pm$0.039} \\

\midrule
\textbf{0.325} &
\cellcolor{ggreen!20}\makecell{100\%\\R:0~~B:0\\0.034$\pm$0.004} &
\cellcolor{ggreen!20}\makecell{100\%\\R:0~~B:0\\0.034$\pm$0.005} &
\cellcolor{ggreen!20}\makecell{100\%\\R:0~~B:0\\0.036$\pm$0.003} &
\cellcolor{ggreen!20}\makecell{100\%\\R:0~~B:0\\0.038$\pm$0.005} &
\cellcolor{ggreen!20}\makecell{100\%\\R:0~~B:0\\0.041$\pm$0.006} &
\cellcolor{ggreen!20}\makecell{92\%\\R:5~~B:3\\0.050$\pm$0.029} &
\cellcolor{ggreen!20}\makecell{89\%\\R:11~~B:0\\0.053$\pm$0.027} &
\cellcolor{gyellow!20}\makecell{74\%\\R:22~~B:4\\0.068$\pm$0.042} &
\cellcolor{gred!20}\makecell{30\%\\R:54~~B:16\\0.104$\pm$0.045} &
\cellcolor{gred!20}\makecell{6\%\\R:80~~B:14\\0.115$\pm$0.038} \\

\midrule
\textbf{0.350} &
\cellcolor{ggreen!20}\makecell{100\%\\R:0~~B:0\\0.039$\pm$0.010} &
\cellcolor{ggreen!20}\makecell{100\%\\R:0~~B:0\\0.038$\pm$0.011} &
\cellcolor{ggreen!20}\makecell{100\%\\R:0~~B:0\\0.041$\pm$0.018} &
\cellcolor{ggreen!20}\makecell{100\%\\R:0~~B:0\\0.040$\pm$0.005} &
\cellcolor{ggreen!20}\makecell{99\%\\R:1~~B:0\\0.045$\pm$0.022} &
\cellcolor{ggreen!20}\makecell{100\%\\R:0~~B:0\\0.045$\pm$0.016} &
\cellcolor{ggreen!20}\makecell{96\%\\R:2~~B:2\\0.050$\pm$0.022} &
\cellcolor{ggreen!20}\makecell{89\%\\R:7~~B:4\\0.066$\pm$0.056} &
\cellcolor{gyellow!20}\makecell{47\%\\R:34~~B:19\\0.106$\pm$0.060} &
\cellcolor{gred!20}\makecell{19\%\\R:58~~B:23\\0.136$\pm$0.057} \\

\midrule
\textbf{0.375} &
\cellcolor{ggreen!20}\makecell{100\%\\R:0~~B:0\\0.049$\pm$0.012} &
\cellcolor{ggreen!20}\makecell{100\%\\R:0~~B:0\\0.052$\pm$0.022} &
\cellcolor{ggreen!20}\makecell{100\%\\R:0~~B:0\\0.048$\pm$0.010} &
\cellcolor{ggreen!20}\makecell{100\%\\R:0~~B:0\\0.048$\pm$0.011} &
\cellcolor{ggreen!20}\makecell{100\%\\R:0~~B:0\\0.052$\pm$0.010} &
\cellcolor{ggreen!20}\makecell{99\%\\R:0~~B:1\\0.054$\pm$0.017} &
\cellcolor{ggreen!20}\makecell{86\%\\R:4~~B:10\\0.058$\pm$0.019} &
\cellcolor{ggreen!20}\makecell{88\%\\R:8~~B:4\\0.079$\pm$0.100} &
\cellcolor{ggreen!20}\makecell{87\%\\R:6~~B:7\\0.087$\pm$0.112} &
\cellcolor{gyellow!20}\makecell{46\%\\R:23~~B:31\\0.113$\pm$0.086} \\

\bottomrule
\end{tabular}

}
    \caption{\textbf{Traversing Straight Stairs.} Each cell reports success rate,
    riser-contact falls (R) and balance falls (B), and mean\,$\pm$\,std 3D
    foot placement error (m). 100 evaluations with initial robot yaw sampled from $\mathcal{U}(-\pi/6,\pi,6)$.}
    \label{tab:apx-stairs}
\end{table*}

\begin{table*}[h]
    \centering
    \resizebox{.8\linewidth}{!}{\setlength{\tabcolsep}{3pt}
\renewcommand{\arraystretch}{1.3}
\tiny
\begin{tabular}{c cccccccccc}
\toprule
\multirow{2}{*}{\textbf{\makecell{Step\\Length\\(m)}}}
  & \multicolumn{10}{c}{\textbf{Step Height (m)}} \\
\cmidrule(lr){2-11}
& \textbf{0.01} & \textbf{0.02} & \textbf{0.04} & \textbf{0.08}
& \textbf{0.10} & \textbf{0.11} & \textbf{0.115}
& \textbf{0.12} & \textbf{0.125} & \textbf{0.13} \\
\midrule

\textbf{0.300} &
\cellcolor{ggreen!20}\makecell{100\%\\R:0~~B:0\\0.057$\pm$0.008} &
\cellcolor{ggreen!20}\makecell{99\%\\R:0~~B:1\\0.053$\pm$0.012} &
\cellcolor{ggreen!20}\makecell{96\%\\R:4~~B:0\\0.049$\pm$0.020} &
\cellcolor{ggreen!20}\makecell{97\%\\R:2~~B:1\\0.045$\pm$0.017} &
\cellcolor{ggreen!20}\makecell{82\%\\R:17~~B:1\\0.061$\pm$0.029} &
\cellcolor{gyellow!20}\makecell{40\%\\R:40~~B:20\\0.082$\pm$0.028} &
\cellcolor{gred!20}\makecell{21\%\\R:61~~B:18\\0.104$\pm$0.031} &
\cellcolor{gred!20}\makecell{14\%\\R:71~~B:15\\0.114$\pm$0.034} &
\cellcolor{gred!20}\makecell{19\%\\R:73~~B:8\\0.112$\pm$0.030} &
\cellcolor{gred!20}\makecell{14\%\\R:72~~B:14\\0.132$\pm$0.053} \\

\midrule
\textbf{0.325} &
\cellcolor{ggreen!20}\makecell{98\%\\R:0~~B:2\\0.066$\pm$0.015} &
\cellcolor{ggreen!20}\makecell{97\%\\R:2~~B:1\\0.066$\pm$0.020} &
\cellcolor{ggreen!20}\makecell{97\%\\R:3~~B:0\\0.061$\pm$0.020} &
\cellcolor{ggreen!20}\makecell{97\%\\R:2~~B:1\\0.049$\pm$0.044} &
\cellcolor{ggreen!20}\makecell{91\%\\R:3~~B:6\\0.060$\pm$0.025} &
\cellcolor{ggreen!20}\makecell{77\%\\R:15~~B:8\\0.066$\pm$0.035} &
\cellcolor{gyellow!20}\makecell{58\%\\R:20~~B:22\\0.085$\pm$0.040} &
\cellcolor{gyellow!20}\makecell{66\%\\R:31~~B:3\\0.082$\pm$0.050} &
\cellcolor{gred!20}\makecell{29\%\\R:59~~B:12\\0.097$\pm$0.038} &
\cellcolor{gred!20}\makecell{17\%\\R:62~~B:21\\0.104$\pm$0.031} \\

\midrule
\textbf{0.350} &
\cellcolor{ggreen!20}\makecell{94\%\\R:2~~B:4\\0.079$\pm$0.039} &
\cellcolor{ggreen!20}\makecell{97\%\\R:1~~B:2\\0.082$\pm$0.030} &
\cellcolor{ggreen!20}\makecell{97\%\\R:1~~B:2\\0.080$\pm$0.029} &
\cellcolor{ggreen!20}\makecell{96\%\\R:0~~B:4\\0.061$\pm$0.068} &
\cellcolor{ggreen!20}\makecell{94\%\\R:4~~B:2\\0.065$\pm$0.088} &
\cellcolor{ggreen!20}\makecell{91\%\\R:3~~B:6\\0.074$\pm$0.098} &
\cellcolor{ggreen!20}\makecell{84\%\\R:8~~B:8\\0.084$\pm$0.095} &
\cellcolor{ggreen!20}\makecell{79\%\\R:13~~B:8\\0.093$\pm$0.107} &
\cellcolor{gyellow!20}\makecell{57\%\\R:25~~B:18\\0.097$\pm$0.059} &
\cellcolor{gred!20}\makecell{28\%\\R:55~~B:17\\0.132$\pm$0.061} \\

\midrule
\textbf{0.375} &
\cellcolor{ggreen!20}\makecell{96\%\\R:1~~B:3\\0.092$\pm$0.044} &
\cellcolor{ggreen!20}\makecell{97\%\\R:2~~B:1\\0.095$\pm$0.051} &
\cellcolor{ggreen!20}\makecell{98\%\\R:1~~B:1\\0.099$\pm$0.026} &
\cellcolor{ggreen!20}\makecell{91\%\\R:4~~B:5\\0.081$\pm$0.038} &
\cellcolor{ggreen!20}\makecell{90\%\\R:4~~B:6\\0.090$\pm$0.137} &
\cellcolor{ggreen!20}\makecell{96\%\\R:1~~B:3\\0.069$\pm$0.048} &
\cellcolor{ggreen!20}\makecell{93\%\\R:1~~B:6\\0.089$\pm$0.126} &
\cellcolor{gyellow!20}\makecell{74\%\\R:7~~B:19\\0.097$\pm$0.067} &
\cellcolor{gyellow!20}\makecell{54\%\\R:16~~B:30\\0.125$\pm$0.107} &
\cellcolor{gred!20}\makecell{17\%\\R:45~~B:37\\0.155$\pm$0.097} \\

\bottomrule
\end{tabular}
}
    \caption{\textbf{Traversing Spiral Stairs.} Each cell reports success rate,
    riser-contact falls (R) and balance falls (B), and mean\,$\pm$\,std 3D
    foot placement error (m). 100 evaluations with initial robot yaw sampled from $\mathcal{U}(-\pi/6,\pi,6)$.}
    \label{tab:apx-spiral}
\end{table*}


\end{document}